\documentclass[11pt]{article}

\usepackage[final]{acl}

\usepackage{times}
\usepackage{latexsym}
\usepackage[T1]{fontenc}
\usepackage[utf8]{inputenc}
\usepackage{microtype}
\IfFileExists{inconsolata.sty}{\usepackage{inconsolata}}{}
\usepackage{graphicx}

\usepackage{subcaption}
\usepackage{enumitem}
\usepackage{listings}
\usepackage{booktabs}
\usepackage{cuted}
\usepackage{caption}

\renewcommand\paragraph[1]{
    \vspace{0.2cm}
    \noindent 
    \textbf{#1}
}

\setlist[itemize]{leftmargin=*, itemsep=2pt, topsep=1pt, parsep=0pt, partopsep=0pt}
\setlist[enumerate]{leftmargin=*, itemsep=2pt, topsep=1pt, parsep=0pt, partopsep=0pt}

\lstdefinestyle{prompt}{
    basicstyle=\ttfamily\footnotesize,
    breaklines=true,
    columns=fullflexible,
    keepspaces=true,
    frame=single,
    aboveskip=4pt,
    belowskip=6pt
}

\title{Limitations of Automated Simulatability:\\LLM Simulators Can Bypass Explanations}

\author{
 \textbf{Antonin Poché\textsuperscript{1,2}},
 \textbf{Fanny Jourdan\textsuperscript{1}},
 \textbf{Nils Feldhus\textsuperscript{3}},
 \textbf{Qianli Wang\textsuperscript{4}},
\\
 \textbf{Jing Yang\textsuperscript{4,8}},
 \textbf{Simon Ostermann\textsuperscript{5,7,9}},
 \textbf{Nicholas Asher\textsuperscript{2,10}},
 \textbf{Philippe Muller\textsuperscript{2}},
 \textbf{Vera Schmitt\textsuperscript{4,6,7,9}},
\\
\\
 \textsuperscript{1}IRT Saint Exupéry,
 \textsuperscript{2}IRIT, Université de Toulouse,
 \textsuperscript{3}University of Groningen,
 \textsuperscript{4}Technische Universit\"at Berlin,\\
 \textsuperscript{5}Saarland University,
 \textsuperscript{6}Johannes Gutenberg-University Mainz,
 {\footnotesize 
 \textsuperscript{7}German Research Center for Artificial Intelligence (DFKI), }\\
 {\footnotesize 
 \textsuperscript{8}BIFOLD – Berlin Institute for the Foundations of Learning and Data,
 \textsuperscript{9}Centre for European Research in Trusted AI (CERTAIN),}
 \textsuperscript{10}CNRS,
\\
 \small{
   \textbf{Correspondence:} \href{mailto:antonin.poche@irt-saintexupery.com}{antonin.poche@irt-saintexupery.com}
 }
}

\begin{document}
\maketitle

% ======== %
% Abstract %
% ======== %
\begin{abstract}
    Simulatability is an evaluation protocol for explanations that quantifies their usefulness by how well they help a user predict a task model's outputs. Since human evaluation is costly, automated simulatability replaces human explainees with LLM simulators, as proposed in ConSim \cite{poche2025consim} for large-scale experiments. We qualitatively replicate and extend ConSim's ranking of explanation methods across the tested datasets, explanation families, and simulator LLMs, and identify two limitations. First, when class names are meaningful, simulators can obtain high simulatability by solving the classification task directly, without relying on the explanations. Second, class anonymization can reward explanations for leaking the hidden label mapping, a limitation we expose with a new classes-as-concepts baseline. These results are consistent with a shortcut hypothesis: in the tested settings, simulator predictions mainly rely on task priors, while explanations produce small changes. We derive recommendations for more robust automated simulatability evaluations.
\end{abstract}

\begin{figure}[!t]
    \centering
    \includegraphics[width=\linewidth]{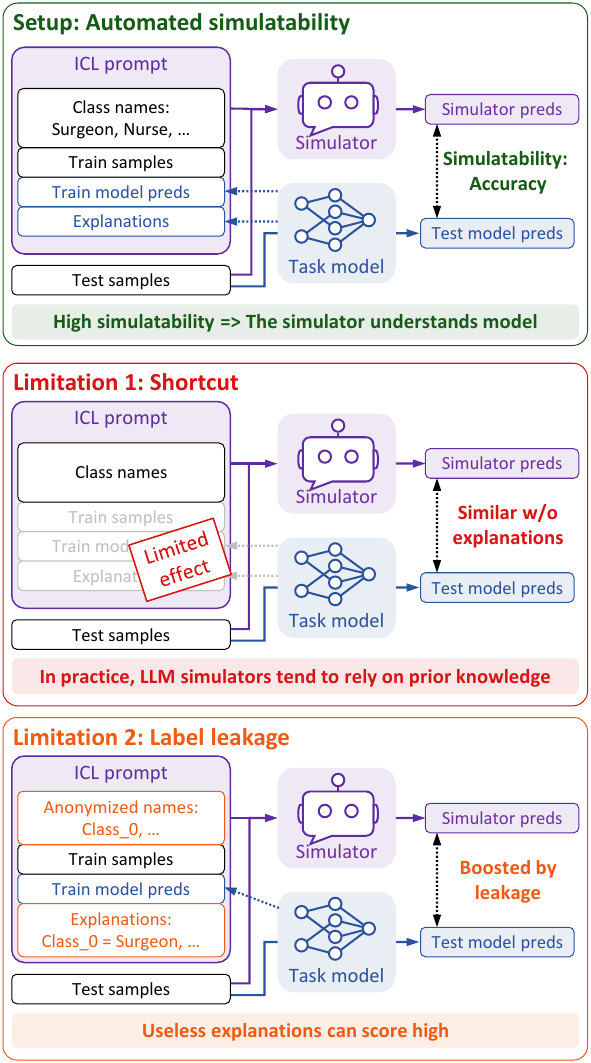}
    \caption{\textbf{Overview of automated simulatability limitations}. Automated simulatability explains how a task model works via ICL (in-context learning), and the LLM simulator must simulate its predictions on new samples. In theory, simulatability measures the usefulness of explanations. However, simulator predictions can remain unaffected by explanations, thus relying on task priors (Sec.~\ref{sec:shortcut}). In anonymized settings, explanation scores can also reward hidden label recovery (Sec.~\ref{sec:anonymized}).}
    \label{fig:main}
    \vspace{-2em}
\end{figure}

% ============ %
% Introduction %
% ============ %
\section{Introduction}

    % Explainability is important
    % There are quantitative metrics for faithfulness and complexity
    % But they do measure how useful explanations are and often miss plausibility
    Explanations of a task model's predictions are often evaluated via faithfulness and complexity metrics \cite{jacovi2020towards}, which only assess internal properties of the explanation, i.e., how well it reflects the task model's decision process or how concise it is, rather than whether the explanation is actually useful or plausible to an explainee.
    
    % Similatability experiments tackle this
    % However, they are costly and complex to set up for significant results
    % So some papers introduced automated simulatability
    Simulatability addresses this gap by evaluating explanations end-to-end: an explanation is useful insofar as it enables an explainee to predict the task model's behavior on held-out inputs \cite{kim2016examples}.
    Early simulatability studies relied on humans in controlled user studies \cite{lage2019human, colin2022cannot} which are costly and difficult to scale. To overcome this limitation, recent work substitutes human explainees with LLM simulators, yielding \textit{automated simulatability} \cite{hase2020evaluating, poche2025consim}.
    
    % We first intended to use it to compare different explanation types to evaluate a new one we wanted to introduce
    % However, in our experiments, no explanation type appeared significantly different to others, including the no-explanations.
    We replicate and extend ConSim's experiments \cite{poche2025consim} to additional explanation types (attributions and rationales), LLM simulators, and datasets. As in the original paper, in non-anonymized experiments, we find that explanations produce small changes to simulator predictions.
    We hypothesize a shortcut: simulator predictions are dominated by task priors rather than explanation content. Echoing similar effects in ICL (in-context learning) \cite{liu2022whatmakes,jang2024rectifying} and, in human psychology, with goal neglect \cite{duncan1996intelligence}. Then we present evidence consistent with this hypothesis.
    To prevent shortcuts, ConSim anonymizes class names; however, we show that this fix has its own limitations. We introduce a classes-as-concepts baseline, which trivially explains a prediction by naming the predicted class, and find that it outperforms every tested method under anonymization, showing that rankings can reward label leakage rather than meaningful explanations. We close with recommendations for robust simulatability protocols.

     % NEW VERSION Following framing - BEGINNING - Vera 
    Our contributions are: (1) a qualitative replication of ConSim's protocol and extension to new datasets, explanation families (attributions and rationales), and LLM simulators; (2) the shortcut hypothesis and cross-family and cross-LLM evidence consistent with strong task-prior influence in non-anonymized settings; (3) the classes-as-concepts baseline, showing that anonymized evaluation can reward uninformative label leakage explanations; (4) concrete recommendations for more robust automated simulatability evaluation.

% ========== %
% Background %
% ========== %
\section{Background}

    % ------------------
    % Explanations types
    \paragraph{Explanations types.}
        Interpretability aims at understanding the behavior and decisions of task models. It can be divided into explanation types; we use three of them (thorough overview in App.~\ref{sec:explanation_types}): \begin{itemize}
            \item \emph{Attributions} are popular methods that quantify the contribution of input features to the prediction \cite{lundberg2017unified,simonyan2014deep}.
            \item \emph{Rationales} are free-form textual justifications of a prediction \cite{camburu2018snli}, most of the time LLM-generated \cite{turpin2023language}.
            \item \emph{Concept-based explanations} map internal task model computations to human-understandable concepts, providing a higher-level view of what a task model has learned \cite{kim2018interpretability, feldhus2025interpreting}. In this paper, we focus on post-hoc unsupervised methods \cite{poeta2023concept}.
        \end{itemize}
    
        Existing metrics largely assess the quality of all these methods with respect to task model behavior or generalization across inputs, rather than whether they are effectively \emph{communicated} to an evaluator.
        This gap motivates \emph{simulatability} as a criterion for evaluating the effectiveness of explanations.

    % --------------
    % Simulatability    
    \paragraph{Simulatability}
        refers to the degree to which explanations help a simulator to predict a task model's outputs \cite{kim2016examples, lage2019human}.
        It shifts the evaluation focus from internal faithfulness to downstream utility: an explanation is useful insofar as it lets an evaluator reproduce the task model's behavior on held-out inputs.
        Early evaluations relied on controlled human user studies \cite{lage2019human, colin2022cannot}, which are difficult to scale. (Deeper related work in App.~\ref{sec:simulatability}).

    % ------------------------
    % Automated Simulatability
    \paragraph{Automated Simulatability.}
        To overcome the scalability ceiling of human studies, recent work substitutes human explainees with LLM simulators \cite{de2024evaluating,nguyen2024interpretable}. However, the simulator's nature changes what is evaluated \cite{chan2022frame}.
        In our paper, we replicate and extend ConSim \cite{poche2025consim}, an automated simulatability framework for concept-based explanations. They notably introduce anonymized experiments (which we detail in Sec.~\ref{sec:anonymized}) to force LLM simulators to rely on the explanations. They find consistent rankings across datasets, task models, and simulators.

% =================================== %
% Replication and extension of ConSim %
% =================================== %
\section{Replication and extension of ConSim} \label{sec:reproduction}

    Before analyzing the limitations, we verify that our implementation recovers the qualitative findings of ConSim \cite{poche2025consim}. We then select a prompt format and concept interpretation for the analyses in Sections \ref{sec:shortcut} and \ref{sec:anonymized}.

    % -----------------
    % Replication setup
    \subsection{Replication setup}
        We detail ConSim experiments in App.~\ref{sec:consim}, but we invite the reader to refer to the original paper for the complete description \cite{poche2025consim}. We call our replication \texttt{old\_consim}.
        
        \paragraph{What is preserved:}
            We keep the same simulatability logic, prompt-type families, and anonymized variants. We use the same methods (NoProjection is renamed neurons-as-concepts), and keep the TopK concept interpretation. We keep 3 of the four datasets: BIOS, IMDB, and Rotten Tomatoes. We also use deterministic sampling to balance the number of correct and incorrect task model predictions.

        \paragraph{What changes:}\begin{itemize}
            \item Task models: We use one HuggingFace task model per dataset rather than retrain multiple architectures. Because their activations are not necessarily positive, we replace NMF with Semi-NMF \cite{ding2010convex}.
            \item Evaluation stability: We use 50 seeds and split them into 10 disjoint groups of 5. Each grouped observation, therefore, aggregates 100 evaluation predictions rather than the 20 available from a single seed, reducing the discreteness and variance of the accuracy estimates.
            \item Coverage: We add AG News~\cite{zhang2015character} and GoEmotions~\cite{demszky2020goemotions}, replace Tweet Eval Emotion \cite{mohammad2018semeval} by Emotion \cite{saravia-etal-2018-carer}, and treat BIOS slightly differently (App.~\ref{sec:class_subset_construction}).
            \item Prompting: We reconstruct the original prompt setup as \texttt{old\_consim} and introduce \texttt{new\_consim}, which includes 3 changes: related information is interleaved with examples, importance values are verbalized, and evaluation samples are predicted one at a time. Details in App.~\ref{sec:consim_prompt_formats}.
            \item Concept interpretation: We also test an LLM-based approach to concept interpretation, addressing one of ConSim's announced limitations.
        \end{itemize}

    % ----------------------------------
    % Implementation and Reproducibility
    \subsection{Implementation and Reproducibility}
        Code, prompts, task model identifiers, seeds, outputs, and reproduction commands are available on \href{https://github.com/AntoninPoche/simulatability_shortcut/tree/clean}{GitHub}. We rely on the library Interpreto \cite{poche2026interpreto} for attribution and concept-based explanations. Interpreto concept-based explanations use the \texttt{LanguageModel} from NNsight \cite{fiotto2025nnsight} for model splitting and activation extraction, and overcomplete \cite{fel2025overcomplete} for concept learning. The \texttt{old\_consim} implementation was available in Interpreto. Finally, plots rely on the code released with the original ConSim paper.

    % -----------------------------------
    % Reproduction and protocol selection
    \subsection{Reproduction and protocol selection}

        % Qualitative reproduction
        \paragraph{Qualitative reproduction.}
            We first assess whether our reconstruction of the original \texttt{old\_consim} protocol recovers the qualitative findings reported by ConSim. Exact numerical replication is not expected because we use different task models, a different simulator, and a partially different set of datasets. We therefore evaluate whether the broad relative ordering of the explanation methods is preserved.
        
            Figure~\ref{fig:consim_copy_pairwise} is a copy of figures reported in the original paper, while Fig.~\ref{fig:old_pairwise} shows our reconstruction. The main pattern is preserved: Vanilla SAE, ICA, and Semi-NMF rank above the no-explanation baseline, whereas PCA and SVD remain near the bottom. Some method-level positions differ, but we do not interpret these differences because the task models and decomposition setup have changed. Overall, our reproduction aligns with the original ConSim study.

        % Prompt-format selection
        \paragraph{Prompt-format selection.}
            We next compare the reconstructed \texttt{old\_consim} format with our revised \texttt{new\_consim} format. The paired results are reported in Fig.~\ref{fig:new_old_diff} in App.~\ref{sec:protocol_comparison}. \texttt{new\_consim} yields significantly higher simulatability scores for 8/10 prompt types. Nonetheless, the rankings of the two formats remain correlated (Pearson $r=.87$ and Spearman $\rho=.87$). Thus, \texttt{new\_consim} generally increases scores while broadly preserving the original behavior. We use \texttt{new\_consim} as the common prompt format in the following experiments. Note that on Qwen-3.5-9B, later conclusions hold with the \texttt{old\_consim} format and LLM-generated interpretations.

        % Concept-interpretation selection
        \paragraph{Concept-interpretation selection.}
            Using \texttt{new\_consim}, we also compare TopK and LLM-generated concept interpretations. TopK yields significantly higher scores for 5/6 applicable concept prompt types (Fig.~\ref{fig:llm_TopK_diff} in App.~\ref{sec:protocol_comparison}). We therefore use TopK as the default concept interpretation in the subsequent analyses.

        % Motivating observation
        \paragraph{Motivating observation.}
            With both \texttt{old\_consim} and \texttt{new\_consim}, the non-anonymized score distributions overlap strongly across concept methods and their matched no-explanation baselines (Fig.~\ref{fig:old_new_TopK_violins}). The next section explores this limitation.

% ======================================================= %
% Task-solving shortcuts in non-anonymized simulatability %
% ======================================================= %
\section{Task-solving shortcuts in non-anonymized simulatability} \label{sec:shortcut}

    The previous section showed that, when class names are visible, concept explanations produce only limited separation from their matched no-explanation baselines. We first test whether this behavior recurs across the tested explanation families and LLM simulators. We then examine the hypothesis that simulators primarily rely on their prior knowledge of the classification task, with only minor changes from the provided explanations.

    % -------------------------------------
    % Attributions and rationales extension
    \subsection{Attributions and rationales extension}

        We extend the non-anonymized evaluation to concept explanations (C), feature attributions (A), and natural-language rationales (R) while keeping the simulator task and prompt structure fixed. All prompts include the task description and evaluation sample; they differ only in whether they contain global explanations, learning examples with task model predictions, and local explanations. Table~\ref{tab:prompt_types} summarizes the resulting prompt types.

        \begin{table}[t]
            \centering
            \resizebox{\columnwidth}{!}{%
                \begin{tabular}{lccccccc}
                    \toprule
                    \textbf{Prompt element} & \textbf{B1} & \textbf{B2} & \textbf{C1} & \textbf{C2} & \textbf{C3} & \textbf{A} & \textbf{R} \\
                    \midrule
                    Task description        & X  & X  & X  & X  & X  & X  & X  \\
                    Global explanations     & -- & -- & X  & X  & X  & -- & -- \\
                    Examples and predictions& -- & X  & -- & X  & X  & X  & X  \\
                    Local explanations      & -- & -- & -- & -- & X  & X  & X  \\
                    \bottomrule
                \end{tabular}
            }
            \caption{Prompt types used for the cross-family shortcut experiments. \textbf{B1} and \textbf{B2} are no-explanation baselines. \textbf{C1--C3} are concept prompts. \textbf{A} is the attribution prompt, where local explanations are word attributions for the learning examples. \textbf{R} is the rationale prompt, where local explanations are natural-language rationales for the learning examples. The evaluation phase never includes an explanation or a label.}
            \label{tab:prompt_types}
            \vspace{-1em}
        \end{table}

        Following Sec.~\ref{sec:reproduction}, concept prompts use \texttt{new\_consim} with TopK interpretations. We select one representative method per explanation family based on the Qwen-3.5-9B \cite{qwen35blog} rankings reported in App.~\ref{sec:methods_ranking}: Vanilla SAE for concepts, LIME for attributions, and Qwen3.5-2B for rationales. We first evaluate the complete grid with Qwen-3.5-9B, then repeat the representative-method comparison with Llama-3.1-8B \cite{dubey2024llama}, Gemma-4-12B \cite{gemmateam2026gemma4technicalreport}, and Phi-4 \cite{abdin2024phi4technicalreport}.

    % -------------------------------------------------------
    % Consistent but small explanation gains across LLM simulators
    \subsection{Consistent but small explanation gains across LLM simulators}
        
        \begin{figure}[t]
            \centering
            \includegraphics[width=\linewidth]{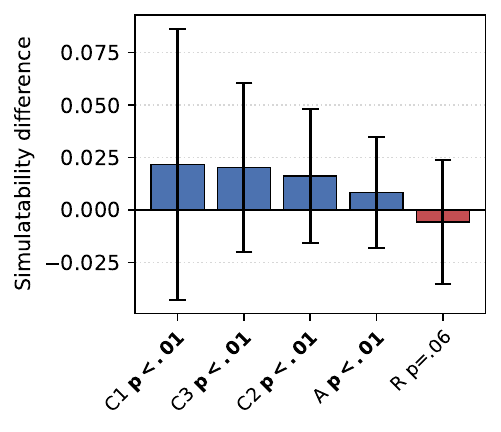}
            \vspace{-2em}
            \caption{\textbf{Qwen-3.5-9B prompt-type differences.} Bar-plot comparing each selected prompt type to its matching no-explanation baseline.}
            \label{fig:families_prompt_type_diff_qwen}
            \vspace{-1em}
        \end{figure}

        With Qwen-3.5-9B, concept and attribution prompts are significantly above their matched no-explanation baselines (Fig.~\ref{fig:families_prompt_type_diff_qwen}). However, the mean gains are small relative to the variability across experimental settings.

        Gemma-4-12B and Phi-4 show similar rankings and conclusions (App.~\ref{sec:llm_judges_barplots}): Explanations provide small gains over their matched baselines, with concept prompts showing the clearest improvements. This consistency is also visible at the prediction level: Qwen-3.5-9B, Gemma-4-12B, and Phi-4 agree on about 85\% of their predictions (Fig.~\ref{fig:prediction_exact_agreement_summaries}).
        
        Three of four retained user-LLMs show similar small gains in explanation. Llama-3.1-8B reverses this pattern, demonstrating that automated simulatability remains simulator-dependent. Its lower agreement with the task model suggests that simulator capability may contribute to this difference.

    % ----------
    % Hypothesis
    \subsection{The shortcut hypothesis}
        A small explanation can admit at least two interpretations. First, the tested explanations may provide little useful information beyond what the LLM simulator already knows. Second, the simulator may not use the explanations even when they contain relevant information, because directly predicting the gold label is easier than inferring the task model's behavior.

        The consistency of the pattern across three explanation families and three LLM simulators makes a shared limitation of the evaluation protocol more plausible than independent failures of each explanation method, although the latter cannot be ruled out. We therefore investigate the following shortcut hypothesis: LLM simulator predictions are mainly determined by task priors, with explanations producing smaller changes.

    % ------------------------------------
    % Evidence for the shortcut hypothesis
    \subsection{Evidence for the shortcut hypothesis} \label{sec:shortcut_evidence}

        \begin{figure}[t]
            \centering
            \includegraphics[width=\linewidth]{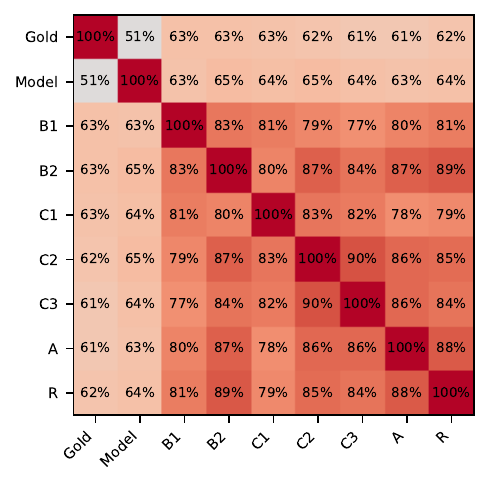}
            \vspace{-2em}
            \caption{\textbf{Prompt types mostly agree.} Mean pairwise exact agreement between gold labels, task model predictions, and selected prompt-type predictions. Each cell is computed within a LLM simulator/dataset/class-subset triplet and averaged equally over all triplets.}
            \label{fig:prediction_exact_agreement_summary}
            \vspace{-1em}
        \end{figure}

        \paragraph{High agreement without learning information.}
            The B1 prompt provides the task description and class names but no learning examples, task model predictions, or explanations (Tab.~\ref{tab:prompt_types}). Nevertheless, B1 achieves simulatability scores above $0.7$ across several datasets (penultimate violin in each group of Fig.~\ref{fig:families_violins_by_judge}). Because evaluation samples are balanced between correct and incorrect task model predictions, always predicting the gold label would yield an expected score of $0.5$. A B1 score of $0.7$, therefore, requires agreement with at least 40\% of the task model's errors. Thus, high agreement can arise without learning examples, task model predictions, or explanations.

        \paragraph{Prompt types mostly agree.}
            Fig.~\ref{fig:prediction_exact_agreement_summary} reports exact agreement between predictions produced under the different prompt types. It shows strong agreement across prompt types, including the B1 (class-names-only) baseline. Therefore, in many cases, including examples or explanations does not change the simulator predictions. We observe even stronger agreement between prompts with learning examples; however, as shown in Fig.~\ref{fig:families_violins_by_judge}, this does not meaningfully change scores.

            In addition, the dataset- and simulator-wise results in App.~\ref{sec:prediction_exact_agreement} show that predictions match the gold label on IMDB nearly 100\% of the time across all prompt types, consistent with a strong reliance on prior knowledge.

\begin{figure}[t]
            \centering
            \includegraphics[width=\linewidth]{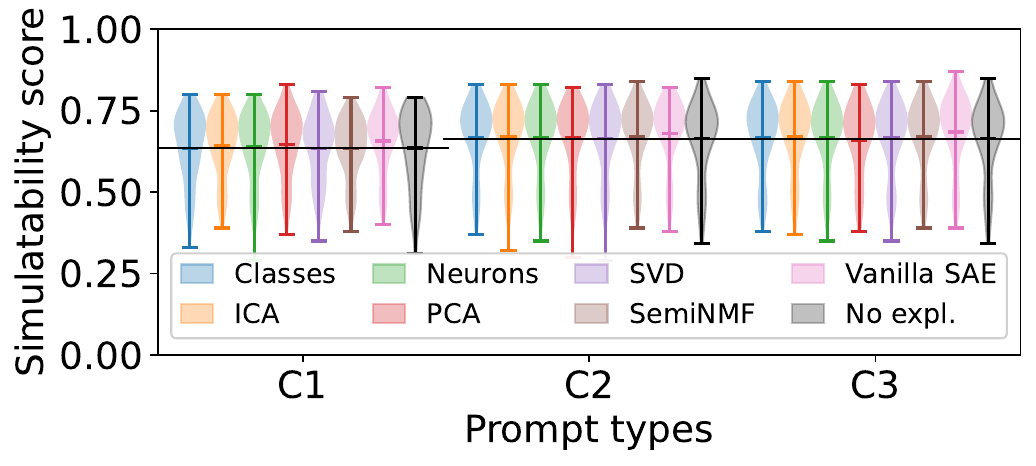}
            \vspace{-2em}
            \caption{\textbf{Non-anonymized concept prompts overlap.} Qwen-3.5-9B grouped-seed scores for TopK \texttt{new\_consim} C1--C3 prompts. The methods include qualitatively distinct concept decompositions, but their distributions largely overlap.}
            \vspace{-1em}
            \label{fig:non_anon_concept_violins}
        \end{figure}

        \paragraph{Qualitatively different concepts have similar scores.}
            The limited score separation is not solely due to equally uninformative concept descriptions. For the BIOS subset \{dentist, physician, surgeon\}, SVD's displayed global concept for \emph{dentist} is ``Scholarship, Ph.D., Professor, Faculty, \ldots'' and is opposed to the class, whereas Vanilla SAE displays the supportive descriptor ``D.D.S., Dentistry, teeth, board-certified, \ldots'' (Table~\ref{tab:qualitative_concept_examples}). Yet, over C1--C3, their mean scores on this subset are equal ($.712$ each). More broadly, Fig.~\ref{fig:non_anon_concept_violins} shows strongly overlapping non-anonymized score distributions for all concept methods, despite such qualitative differences.

        \paragraph{The harder the classification task, the clearer the differences.}
            The clearest explanation gains occur on the selected GoEmotions subsets, which contain semantically confusable labels such as anger/annoyance/neutral and admiration/approval/caring. The task model's F1 score on this dataset is $0.541$ (HuggingFace model id: \texttt{SamLowe/roberta-base-go\_emotions}). These subsets also have lower no-explanation performance, leaving more room for learning-phase information to help. This dataset dependence is consistent with the shortcut hypothesis: task priors dominate when the classification problem is easy, whereas explanation matters more when direct prediction is difficult.

        \paragraph{Explicit simulatability framing does not improve agreement.}
            Finally, we compare \texttt{new\_consim} with \texttt{simulator\_consim}, which explicitly instructs the LLM simulator to simulate the task model. All other prompt content and experimental settings are held fixed (App.~\ref{sec:consim_prompt_formats}). Mentioning the task model generally reduces simulatability scores (Fig.~\ref{fig:sim_new_diff}). Thus, explicit simulator framing alone does not align with task model behavior.

        These observations are consistent with the shortcut hypothesis: when direct classification is easy, simulator predictions change little when explanations are added. They do not establish that explanations are unused.

        ConSim forces the simulator to rely on explanations by anonymizing the output classes. The next section examines whether anonymization successfully removes the shortcut.

% ======================================= %
% Anonymization creates a second shortcut %
% ======================================= %
\section{Anonymization sensitivity to leakage} \label{sec:anonymized}

    In theory, anonymization from ConSim \cite{poche2025consim} prevents LLM simulators from directly predicting meaningful class names. However, we show that explanations can reveal the hidden label mapping, creating a second shortcut.

    % ----------------------
    % Anonymized experiments
    \subsection{Anonymized experiments}

        Anonymized experiments replace class names with generic labels such as \texttt{Class\_1} and \texttt{Class\_2}. The simulator must predict these labels during evaluation. This removes the direct semantic link between the input and the expected output, preventing the simulator from solving the classification task solely using class names. Instead, it must infer the label mapping from the learning-phase examples and explanations. Without learning-phase information, the anonymized B1 baseline is expected to perform at the chance level.

    % -------------------
    % Classes as concepts
    \subsection{Classes-as-concepts}

        Anonymization hides the output labels but not the semantic information contained in explanations. An explanation may therefore reveal which original class each anonymous label corresponds to.

        To demonstrate this failure mode, we introduce the \textbf{classes-as-concepts} baseline, where each concept corresponds directly to one class. In a non-anonymized prompt, this produces redundant explanations, such as associating the prediction ``surgeon'' with the concept ``surgeon.'' In an anonymized prompt, however, it associates \texttt{Class\_0} with ``surgeon,'' directly revealing the hidden label mapping.

\begin{figure}
            \centering
            \includegraphics[width=\linewidth]{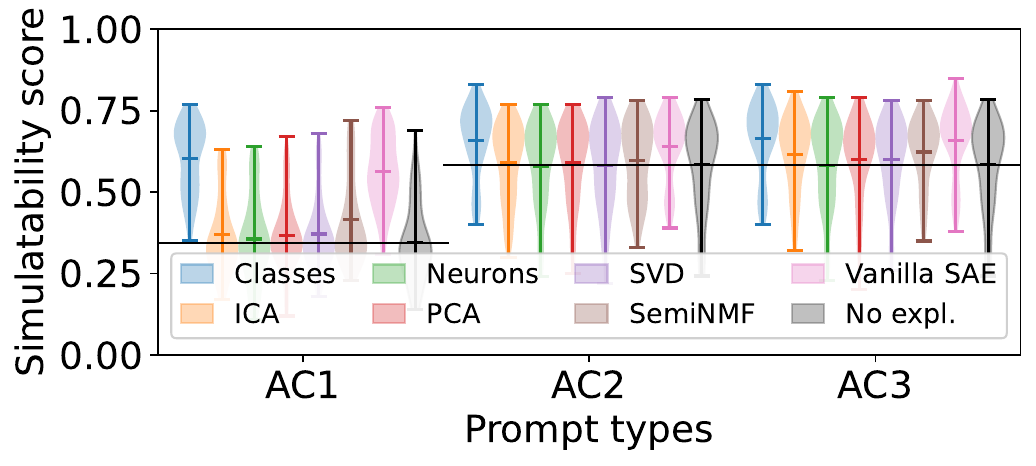}
            \vspace{-2em}
            \caption{\textbf{Classes-as-concepts in anonymized concept prompts.} Violin plot on Qwen-3.5-9B concept scores with TopK \texttt{new\_consim} and anonymized prompt types.}
            \vspace{-1em}
            \label{fig:classes_as_concepts}
        \end{figure}

        Figure~\ref{fig:classes_as_concepts} shows that classes-as-concepts obtain the highest scores across all anonymized prompt types. The pairwise comparisons in Fig.~\ref{fig:new_TopK_anon_pairwise} confirm that they outperform all other concept methods. This ranking does not indicate a better explanation of task model behavior: the baseline succeeds by revealing the mapping between anonymous labels and class names.

        Anonymizing the explanation strings would not fully solve this issue, since synonyms or other semantic descriptions could reveal the same mapping.

        Anonymized simulatability can therefore reward label leakage rather than explanation quality. We recommend including classes-as-concepts, or a similar label-leakage baseline, whenever anonymized prompts are used.

% =============================== %
% Conclusions and recommendations %
% =============================== %
\section{Conclusions and recommendations}

    % Conclusions
    \paragraph{Conclusions.}
        In this paper, we introduced a prompting protocol and qualitatively replicated the ConSim ordering. As in the ConSim paper, we observe that non-anonymized explanations provide limited separation from no-explanation baselines.

        We show that this pattern generalizes across concepts, attributions, and rationales and LLM simulators. Although explanation gains are statistically significant, they remain small relative to their variability. We propose the shortcut hypothesis to explain this result: simulator predictions are often dominated by direct task solving, with explanations producing only secondary changes. We present several observations consistent with this hypothesis.

        ConSim uses class anonymization to prevent direct task solving. However, we find that anonymized evaluations can instead reward explanations that solely reveal the hidden label mapping. We demonstrate this failure mode with the classes-as-concepts baseline, which outperforms the other concept methods under anonymization despite providing no information on the task model.

        Both limitations show that automated simulatability scores can be misleading. These findings lead to the following recommendations:
    
    % Recommendations
    \paragraph{Recommendations:}
        \begin{itemize}
            \item Over-sample the task model errors to emphasize model-specific behavior. It spreads scores and makes them more interpretable.
            \item Include a label leakage baseline, ensuring the ranking favors meaningful explanations (Sec.~\ref{sec:anonymized})
            \item Use multiple LLM simulators, report simulator-specific results, and favor better LLMs.
            \item Use enough seeds for stable estimates, and report effect sizes and statistical significance.
            \item Use novel or semantically difficult tasks that cannot be solved easily from pretrained task knowledge, reducing shortcuts opportunities (Sec.~\ref{sec:shortcut}).
        \end{itemize}

% =========== %
% Limitations %
% =========== %
\section*{Limitations}

    Our findings use an LLM simulator with up to 15B parameters without reasoning. They might not generalize to larger, better, or closed-source LLMs.

    We focused our analysis on use cases that LLM simulators have surely already encountered during training or that they have sufficient semantic knowledge to solve. In other words, we did not work around contamination \cite{balloccu-etal-2024-leak}.

% ================= %
% Ethical statement %
% ================= %
\section*{Ethical statement}

    Coding AI assistants were used throughout the project to help debug, help scale the experiments on a cluster (we used 590 H100 hours), improve visualizations, and detail appendices. Writing AI assistants were mainly used to simulate reviews to improve the general coherence and narrative flow. Everything was human-reviewed at least once.

% =============== %
% Ackowledgements %
% =============== %
\section*{Ackowledgements}

    Our work has benefited from the AI Cluster ANITI and the research programs DEEL\footnote{\url{https://www.deel.ai/}} and FOR\footnote{\url{https://www.irt-saintexupery.com/for-program/}}. ANITI is funded by the France 2030 program under the Grant agreement n°ANR-23-IACL-0002. DEEL and FOR are integrative programs of the AI Cluster ANITI, designed and operated jointly with IRT Saint Exupéry, with the financial support from its industrial and academic partners and the France 2030 program under the Grant agreement n°ANR-10-AIRT-01.
    
    This research is also supported by the National Research Foundation, Prime Minister’s Office, Singapore under its Campus for Research Excellence and Technological Enterprise (CREATE) programme.

    This work was granted access to the HPC/AI resources of IDRIS under the allocation AD011017304 made by GENCI. Granting access to an H100 partition on the Jean Zay super-cluster.

    Finally, Antonin Poché thanks the entire XplaiNLP group for their warm welcome at TU Berlin during the research stay that led to this paper. This research supports the research project FakeXplain by the Berlin Institute for the Foundations of Learning and Data (BIFOLD) and the Federal Ministry of Education and Research (BMFTR) in the scope of the research project VeraXtract (ref. 16IS24066).

% ============ %
% Bibliography %
% ============ %
\bibliography{biblio}

\clearpage
\appendix

\section{Explanations types} \label{sec:explanation_types}

    Interpretability aims at understanding the behavior and decisions of task models. It can be divided into different explanation types (also called explanation families \cite{poche2023natural}):

    % Attributions
    \paragraph{Attributions}
        are popular methods for quantifying the contribution of input features (pixels, tokens, etc.) to the prediction. Here we use two types of such methods, perturbation-based \cite{zeiler2014visualizing,ribeiro2016should,lundberg2017unified,fel2021sobol}, 
        \cite{lundberg2017unified},
        and gradient-based \cite{simonyan2014deep,shrikumar2017learning,smilkov2017smoothgrad,sundararajan2017axiomatic,hooker2018benchmark}. 
        We do not include architecture-based \cite{springenberg2014striving,bach2015pixel,selvaraju2017grad} or attention-based \cite{achtibat2024attnlrp}.

    % Rationales
    \paragraph{Rationales}
        Another explanation type is natural-language rationales, which are free-form textual justifications of a prediction \cite{camburu2018snli,rajani2019explain,narang2020wt5,wang-etal-2025-cross}. They are commonly generated through self-rationalization or elicited as chains of thought \cite{kojima2022large,wei2022chain}. They can seem plausible to humans, while not faithful to the task model \cite{turpin2023language,jacovi2020towards,atanasova2023faithfulness,lanham2023measuring,turpin2023language, sun2026investigatinginterplaycontextualparametric}.

    % Concept-based explanations
    \paragraph{Concept-based explanations}
        map internal task model computations to human-understandable concepts, providing a higher-level view of what a task model has learned \cite{kim2018interpretability, feldhus2025interpreting}. In this paper, we focus on post-hoc unsupervised methods \cite{poeta2023concept}, 
        that explain an already trained task model without prior knowledge of its concepts. This paradigm follows a three-stage pipeline \cite{fel2023holistic,poche2025consim}: \begin{enumerate}
            \item \emph{Concepts learning}: Split the task model into two (in our case, between the encoder and the classification), then hidden representations (here [CLS] tokens) 
            are factorized into basis directions using dictionary learning. (PCA, NMF, SAE...) \cite{fel2023holistic}.
            \item \emph{Concepts' interpretation}: abstract obtained directions are assigned a semantic e.g. via top-$k$ activating tokens \cite{dalvi2022discovering,geva2022transformer} 
            or LLM-generated labels. \cite{bills2023language}.
            \item \emph{Concepts' importance}: 
            Not all concepts are used in a given prediction or are important for a class. 
            concepts' contributions toward the predictions (local) or classes (global) are weighted using gradient-input \cite{shrikumar2017learning} 
            attribution methods on the concepts-to-output function.
        \end{enumerate}

    % Concepts evaluations
    \paragraph{Concepts evaluations}
        A growing body of work investigates what makes concept spaces meaningful and how to evaluate them.
        \citet{geva2022transformer} shows that transformer feed-forward layers naturally promote discrete, vocabulary-level concepts, providing empirical grounding for the idea that neural networks learn concept-like representations.
        Completeness-aware approaches \cite{yeh2020completeness} argue that concept-based explanations should cover the full space of relevant concepts rather than cherry-picking the most salient ones.
        CEBaB \cite{abraham2022cebab} estimates the causal effects of real-world concepts on NLP task model behavior through intervention-based evaluation.
        \citet{bhalla2024towards} propose a unifying framework for evaluating interpretability methods via intervention, while RAVEL \cite{huang2024ravel} benchmarks concept discovery methods on their ability to disentangle language model representations.
        \citet{zarlenga2023towards} propose robust metrics for evaluating concept representation, focusing on stability and faithfulness under perturbations.

\section{Simulatability details} \label{sec:simulatability}

    Simulatability refers to the degree to which an explainee can correctly predict a task model's outputs when provided with its explanations \cite{kim2016examples, lage2019human}.
    It shifts the evaluation focus from internal faithfulness to downstream utility: an explanation is useful insofar as it lets an evaluator reproduce the task model's behavior on held-out inputs.
    The DARPA XAI program formalized this intuition, defining simulatability as the ability of a user to predict a task model's behavior from its explanation \cite{gunning1996examining}.
    Early evaluations relied on controlled human user studies \cite{lage2019human, colin2022cannot}, which are difficult to scale.
    \citet{hase2020evaluating} distinguishes between \emph{forward simulation} (predicting a task model's output from an input and explanation) and \emph{counterfactual simulation} (predicting behavior on a perturbed input after observing the original input, output, and explanation).
    \citet{chen2024models} apply counterfactual simulatability to self-generated natural language explanations, showing that models do not consistently explain themselves on perturbed inputs.
    \citet{limpijankit2025counterfactual} extend counterfactual simulatability to generation tasks, finding that explanations help more for skill-based than knowledge-based tasks.
    \citet{hong2026notall} and \citet{hong2026dollo} compare verbalized feature attributions with self-generated rationales under counterfactual simulation, showing that CoT rationales provide substantially stronger simulation signals than attribution-based explanations.
    \citet{yao2023treu} augment simulatability with a \emph{helpfulness} signal measuring downstream gains at fine-tuning and inference, arguing that simulatability alone falls short for evaluating human-annotated explanations.
    \citet{mills2023almanacs} introduce ALMANACS, a benchmark for evaluating language model explainability through simulatability, covering multiple explanation types and tasks.
    To overcome the scalability ceiling of human studies, recent work substitutes human explainees with LLM simulators.

\section{ConSim experiments details} \label{sec:consim}

    In ConSim \cite{poche2025consim}, the authors proposed an automated simulatability metric for concept-based explanations. The simulator is introduced to the classification task in an initial phase, sees samples and model predictions, and in a learning phase, and is then asked to predict the model's outputs on held-out samples. Their prompts vary in which elements are shown: two no-explanation baselines and three concept-explanation settings, with anonymized variants described in \ref{sec:anonymized}. They then aggregate scores through Copeland-style pairwise comparisons \cite{copeland1951reasonable}, so methods are only compared on matched experimental settings.

    \paragraph{Prompt types}
        They define 5 prompt types, two baselines without explanations (B1 and B2), three concept-based prompts (C1, C2, and C3), and the anonymized versions of these prompts. Prompt-types overview in Tab.~\ref{tab:prompt_types}, details are given in App.~\ref{sec:consim_prompt_types}, and anonymized experiments are detailed in Sec.~\ref{sec:anonymized}.

    \paragraph{Datasets and models}
        The original experiments cover BIOS10 \cite{de2019bias}, IMDB \cite{maas2011learning}, Rotten Tomatoes \cite{pang2005seeing}, and Tweet Eval Emotion \cite{barbieri2020tweeteval}. They evaluate several model families: DistilBERT \cite{sanh2019distilbert}, T5 \cite{raffel2020exploring}, and Llama-3-8B \cite{dubey2024llama}, including positively fine-tuned DistilBERT and T5 variants required by non-negative matrix factorization (NMF).

    \paragraph{Decomposition and interpretation methods}
        The concept extraction methods are NMF \cite{lee1999learning}, Sparse Auto-Encoders (SAE) \cite{ng2011sparse,makhzani2013k}, ICA \cite{hyvarinen2000independent}, PCA \cite{pearson1901liii,hotelling1992relations}, SVD \cite{eckart1936approximation}, and a NoProjection baseline where neurons are treated as concepts. Concepts are communicated either through TopK words \cite{dalvi2022discovering,geva2022transformer} (called CMAW in the original paper and sometimes MaxAct in the literature) or through alignment to existing labels (o1CA). The reported ranking places NMF, SAE, and ICA above others, PCA and SVD are worse than the baselines, while TopK is more reliable than o1CA.

    \paragraph{Samples and seeds}
        Each prompt contains 40 selected samples per seed, split into 20 learning-phase and 20 evaluation-phase samples. The selection is balanced so that half of the samples are correctly predicted by the task model and half are misclassified, making the task about simulating the task model rather than solving the underlying dataset. The main experiments use five random seeds, plus additional seeds for selecting the number of concepts.

\section{Class-subset construction} \label{sec:class_subset_construction}

    \begin{table*}[t]  % Moved for more natural placement of the old and new consim prompt formats
        \centering
        \small
        \begin{tabular}{p{.17\textwidth}p{.23\textwidth}p{.23\textwidth}p{.23\textwidth}}
            \toprule
            \textbf{Property} & \textbf{\texttt{old\_consim}} & \textbf{\texttt{new\_consim}} & \textbf{\texttt{simulator\_consim}} \\
            \midrule
            Task framing & Assign a class to each sample & Assign a class to the evaluation sample & Reproduce the class assigned by another classifier, even when disagreeing \\
            Learning block & Texts, contributions, and predictions in separate blocks & Text, label, and contributions interleaved per sample & Same as new, with ``Model's prediction'' replacing ``Label'' \\
            Importance display & Concept IDs with \texttt{--}/\texttt{-}/\texttt{+}/\texttt{++} & Interpreted concept descriptors with verbal categories & Identical to new \\
            Evaluation requests & One request containing 20 numbered samples & 20 requests containing one sample each & 20 requests containing one sample each \\
            Requested output & 20 lines of \texttt{Sample\_i: class} & One class name & One classifier-predicted class name \\
            \bottomrule
        \end{tabular}
        \caption{Controlled comparison of the three ConSim prompt specifications. New versus old changes prompt organization, importance rendering, and evaluation batching. Simulator versus new changes only task-facing wording and field names.}
        \label{tab:consim_formats}
    \end{table*}

    BIOS \cite{de2019bias} and GoEmotions \cite{demszky2020goemotions} each define 28 labels. Running every simulatability configuration on all 28 labels would too many samples to learn the task model behavior on each class. Furthermore, it would be much more complicated. We therefore run the simulatability evaluation on fixed three-class subsets. Keeping the same cardinality is important as it allows to compare scores between class-subsets and aggregate them. Three classes provided a compromise between retaining confusable decisions and keeping the experiments tractable.
    
    We selected the subsets from confusion matrices between the gold labels and the predictions of the corresponding off-the-shelf task models, 
    \texttt{Fannyjrd/roberta-bios-biased} and \texttt{SamLowe/roberta-base-go\_emotions}. We prioritized groups with large off-diagonal confusion relative to their class frequency, while limiting overlap between groups. Table~\ref{tab:class_subsets} reports the resulting subsets. The two BIOS subsets containing professor are the only overlap among the retained subsets within either dataset.
    
    \begin{table}[t]
        \centering
        \small
        \resizebox{\columnwidth}{!}{%
            \begin{tabular}{lll}
                \toprule
                \textbf{Dataset} & \textbf{Global class IDs} & \textbf{Class names} \\
                \midrule
                BIOS & $[6,19,25]$ & dentist, physician, surgeon \\
                BIOS & $[21,22,26]$ & professor, psychologist, teacher \\
                BIOS & $[1,21,24]$ & architect, professor, software engineer \\
                BIOS & $[9,11,18]$ & filmmaker, journalist, photographer \\
                GoEmotions & $[2,3,27]$ & anger, annoyance, neutral \\
                GoEmotions & $[0,4,5]$ & admiration, approval, caring \\
                \bottomrule
            \end{tabular}%
        }
        \caption{Three-class subsets retained for BIOS and GoEmotions. IDs refer to the original datasets' global label indices.}
        \label{tab:class_subsets}
    \end{table}
    
    The subsets constrain sample selection so that both the label and the task model prediction fall in the subset. For both datasets, concept decompositions are fitted to representations from all 28 classes, and their configured number of concepts is also computed from the full 28-class label space. Thus, a subset experiment evaluates how explanations from a global concept space convey the task model's behavior in a single confusable three-class decision problem.
    
    For each subset and random seed, we cache 40 test samples, of which 20 are used as learning-phase examples and 20 as evaluation samples. Selection is stratified so that, over the complete 40-sample set, half of the task model predictions are correct and half are errors; every class in the subset is represented among both pools whenever the data permit. This balancing makes evaluation accuracy measure agreement with the task model rather than ordinary accuracy against the gold labels. The split and final order are deterministic for a fixed dataset, subset, sample count, and seed. We use 50 seeds for every retained subset.

\section{ConSim prompt formats} \label{sec:consim_prompt_formats}

    We evaluate three prompt specifications. \texttt{old\_consim} reconstructs the organization used in ConSim \cite{poche2025consim}; \texttt{new\_consim} reorganizes the same information and requests one prediction at a time; and \texttt{simulator\_consim} keeps the new organization but explicitly frames the task as reproducing a task model. All three use the same cached task model predictions and the same 40 selected samples for a fixed dataset, class subset, and seed. Consequently, differences between specifications do not come from resampling.

    \subsection{Prompt types and shared inputs} \label{sec:consim_prompt_types}

        The experimental prompt types are listed in Table~\ref{tab:prompt_types}. Prefixing a prompt type with \texttt{A} gives its class-anonymized counterpart: \texttt{AB1}, \texttt{AC1}, \texttt{AB2}, \texttt{AC2}, and \texttt{AC3}. In those prompts, displayed class names and expected answers are replaced by \texttt{Class\_i}. The historical format numbers these labels by their order within the subset, whereas the new and simulator formats retain the datasets' global class IDs. The upper-bound prompt supported by the historical implementation, which exposes local explanations for evaluation samples, is not included in our experiments.
        
        For C1--C3, global explanations associate each displayed class with its most important concepts. C3 additionally associates each learning example with the concepts contributing to the task model's prediction. B1 and C1 contain no learning examples; B2, C2, and C3 contain the same 20 examples and task model predictions. Every type is evaluated on the remaining 20 samples.

    \subsection{Old and new organization}

        Table~\ref{tab:consim_formats} summarizes the structural differences. In \texttt{old\_consim}, the system message first lists all learning texts, then all local-contribution entries when present, and finally all task model predictions. Concept interpretations are listed separately, while global and local importances use concept IDs and the symbols \texttt{--}, \texttt{-}, \texttt{+}, and \texttt{++}. Its single user message contains all 20 evaluation texts, numbered \texttt{Sample\_20} through \texttt{Sample\_39}.
        
        In \texttt{new\_consim}, each learning block instead interleaves the text, task model prediction, and optional local explanation. A concept ID and its interpretation are displayed together, and importance symbols are verbalized as \emph{Very opposed}, \emph{Opposed}, \emph{Supportive}, or \emph{Highly supportive}. The shared system message is paired with 20 separate user messages, each containing one evaluation text. This removes the need for the LLM simulator to maintain a 20-line output protocol and lets every evaluation prediction receive its own generation budget.
        
        The following shortened C3 excerpts preserve the exact field organization while replacing sample text and explanation contents with placeholders:

\noindent\textbf{\texttt{old\_consim}.}
\begin{lstlisting}[style=prompt]
Sample_0: [learning text]
...
Concepts contributions for Sample_0: {1: '+', 3: '++'}
...
Sample_0: pos
...
User: Sample_20: [evaluation text]
Sample_21: [evaluation text]
...
\end{lstlisting}

\noindent\textbf{\texttt{new\_consim}.}
\begin{lstlisting}[style=prompt]
Sample_0:
    Text: [learning text]
    Label: pos
    Concepts contributions: {C3 ([interpretation]): Highly supportive}
...
User: Evaluation sample:
    Text: [one evaluation text]
    Label:
\end{lstlisting}

    \subsection{Explicit simulator framing}
    
        \texttt{simulator\_consim} inherits sample handling, prompt types, concept rendering, and evaluation batching from \texttt{new\_consim}. Its instruction begins: ``You are simulating a text classifier. For the given evaluation sample, predict the class label this classifier would assign.'' It then states: ``Your goal is to reproduce the classifier's output, even when you would otherwise disagree.'' Learning and evaluation fields use \texttt{Model's prediction} instead of \texttt{Label}. No explanations, predictions, or class mappings are otherwise changed.
        
        The simulator grid uses the six datasets, their retained class subsets, all seven concept configurations (Semi-NMF, ICA, PCA, SVD, Vanilla SAE, neurons-as-concepts, and classes-as-concepts), 50 seeds, and the ten standard and anonymized prompt types. It uses TopK interpretations for methods requiring an interpretation; classes-as-concepts directly uses class names and stores no interpretation key. The simulator grid therefore matches the TopK portions of the old and new grids; LLM-generated concept interpretations are not evaluated under simulator framing.

    \subsection{Generation, parsing, and scoring}
    
        Every prompt group has 20 expected task model labels. New and simulator groups produce 20 independent generations, from which one allowed class label is parsed per generation. Old groups produce one generation expected to contain 20 indexed lines.
        
        The old format costs less because all 20 samples are answered together; comparable new-format efficiency would require caching.
        
        For every specification, unparsable outputs are treated as invalid rather than incorrect. A score is reported as the number correct divided by the number valid, only when at least 14 of the 20 expected labels are valid; otherwise, the prompt-group score is undefined. Accuracy is therefore conditional on valid responses, separating formatting failures from disagreement with the task model.

        In practice, less than 1\% of predictions were invalid, so we do not report it.

\section{Protocol comparison} \label{sec:protocol_comparison}

\begin{figure}[!t]
        \centering
        \includegraphics[width=\linewidth]{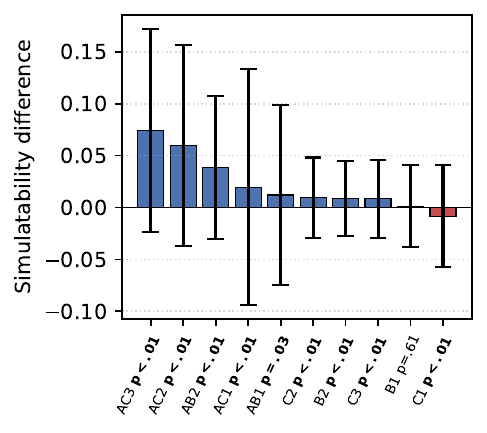}
        \caption{\textbf{New- versus old-ConSim prompt format.} Mean paired Qwen-3.5-9B score differences $\texttt{new\_consim}-\texttt{old\_consim}$. Positive bars favor the revised format.}
        \label{fig:new_old_diff}
    \end{figure}

\begin{figure}[!t]
        \centering
        \includegraphics[width=\linewidth]{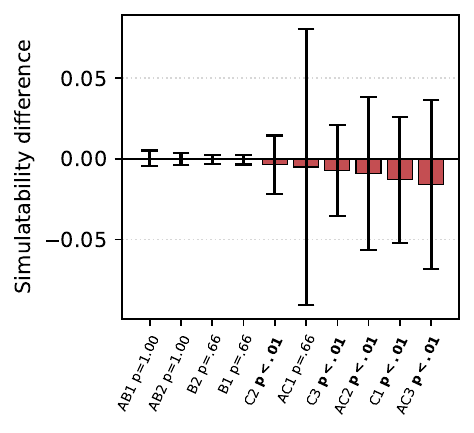}
        \caption{\textbf{LLM-generated versus TopK concept interpretations.} Mean paired Qwen-3.5-9B score differences $\mathrm{LLM}-\mathrm{TopK}$ under \texttt{new\_consim}. Negative bars favor TopK. Only C1--C3 and AC1--AC3 represent applicable concept prompts.}
        \label{fig:llm_TopK_diff}
    \end{figure}

\begin{figure}[!t]
        \centering
        \includegraphics[width=\linewidth]{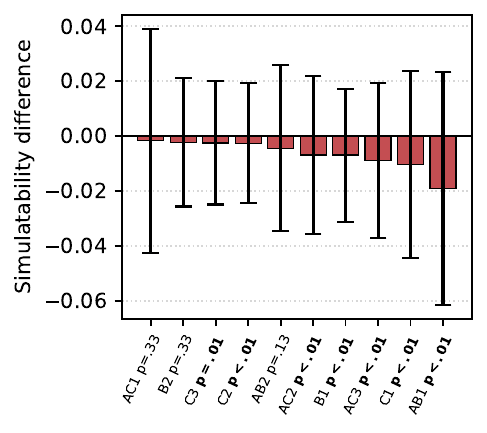}
        \caption{\textbf{Explicit simulator framing versus new-ConSim.} Mean paired Qwen-3.5-9B score differences $\texttt{simulator\_consim}-\texttt{new\_consim}$. Negative bars favor \texttt{new\_consim}.}
        \label{fig:sim_new_diff}
    \end{figure}

    Figures~\ref{fig:new_old_diff}--\ref{fig:sim_new_diff} report paired differences between the prompt and interpretation variants considered in the main text. Scores are matched on all remaining experimental factors, including dataset, task model, class subset, explanation method, grouped seed, prompt type, and, when applicable, concept interpretation. Bars show mean paired score differences over five-seed-grouped observations, error bars show one standard deviation, and tick labels report Holm-corrected two-sided one-sample $t$-tests against zero. Bold tick labels indicate corrected $p<.05$.

    The first comparison contrasts our revised \texttt{new\_consim} prompt format with the reconstructed \texttt{old\_consim} format. Positive values in Fig.~\ref{fig:new_old_diff} indicate higher simulatability under \texttt{new\_consim}. The revised format yields significantly higher scores for eight of the ten prompt types, significantly lower scores for C1, and no significant difference for B1. Despite these score shifts, the two formats remain strongly correlated across matched configurations, as reported in Sec.~\ref{sec:reproduction}. We therefore use \texttt{new\_consim} in the subsequent experiments.

    The second comparison contrasts LLM-generated and TopK concept interpretations under \texttt{new\_consim}. Negative values in Fig.~\ref{fig:llm_TopK_diff} indicate higher scores with TopK. This comparison is only meaningful for concept prompt types C1--C3 and AC1--AC3; baseline rows are included only because their scores are independent of the interpretation choice. TopK yields significantly higher scores for five of the six applicable prompt types, with no significant difference for AC1. We therefore use TopK as the default concept interpretation.
    
    The final comparison contrasts \texttt{simulator\_consim} with \texttt{new\_consim}; the former differs only in explicitly instructing the LLM simulator to reproduce another classifier's predictions. Negative values in Fig.~\ref{fig:sim_new_diff} indicate higher scores under \texttt{new\_consim}. We discuss this comparison as evidence about explicit simulator framing in Sec.~\ref{sec:shortcut_evidence}.
    
    Higher simulatability scores do not necessarily imply higher explanation quality. These comparisons are used to select a common experimental protocol and to study sensitivity to prompt framing, rather than to rank explanation methods.

\clearpage
\section{ConSim reproduction plots with Qwen-3.5-9B judge} \label{sec:consim_reproduction_appendix}

    This appendix details the concept-method comparisons supporting Sec.~\ref{sec:reproduction}. On the following methods: Semi-NMF, ICA, PCA, SVD, Vanilla SAE, neurons-as-concepts, and the no-explanation baseline. The pairwise reproduction matrices also exclude classes-as-concepts because it is introduced as a diagnostic baseline in Sec.~\ref{sec:anonymized}, rather than as a method from the original ConSim comparison.
    
    We average each five consecutive seeds into one of ten grouped-seed observations. For every method pair, we intersect rows on dataset, task model, class subset, grouped seed, and prompt type. A win-rate cell is the proportion of matched rows on which the row method has the higher score, with ties counting as half a win. Difference cells report the mean and standard deviation of the paired score differences. Bold values indicate paired Student $t$-tests \cite{student1908probable} that remain significant at $\alpha=.05$ after Holm correction \cite{holm1979simple} across the 21 unique method pairs displayed in that matrix. Matrix order follows the same Copeland-style \cite{copeland1951reasonable} count of pairwise win rates at least 50\% as the original analysis .

    \paragraph{Copy of figures from the ConSim paper}

\begin{figure*}[!t]
            \centering
            \begin{subfigure}{.49\textwidth}
                \centering
                \includegraphics[width=\linewidth]{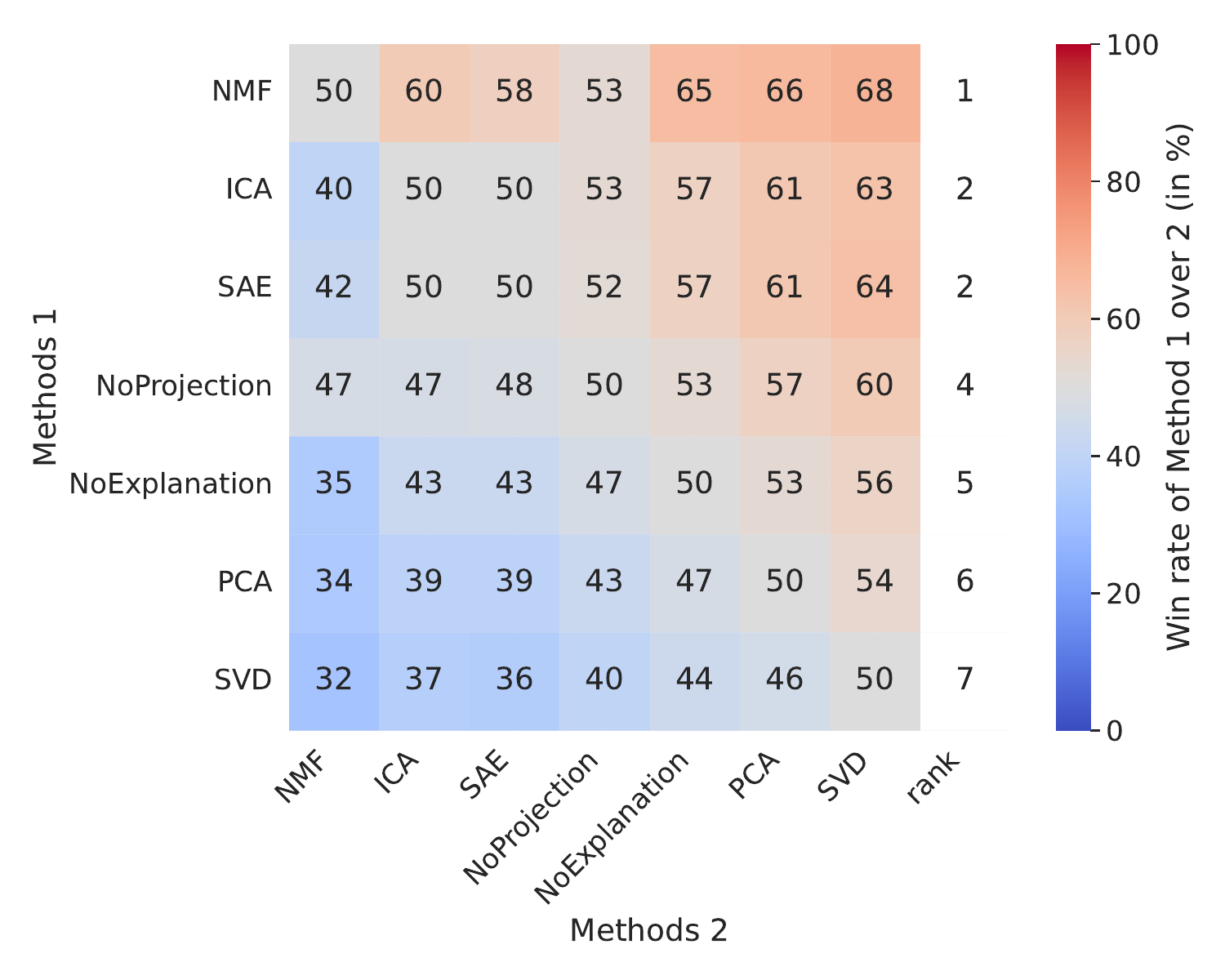}
            \end{subfigure}%
            \begin{subfigure}{.49\textwidth}
                \centering
                \includegraphics[width=\linewidth]{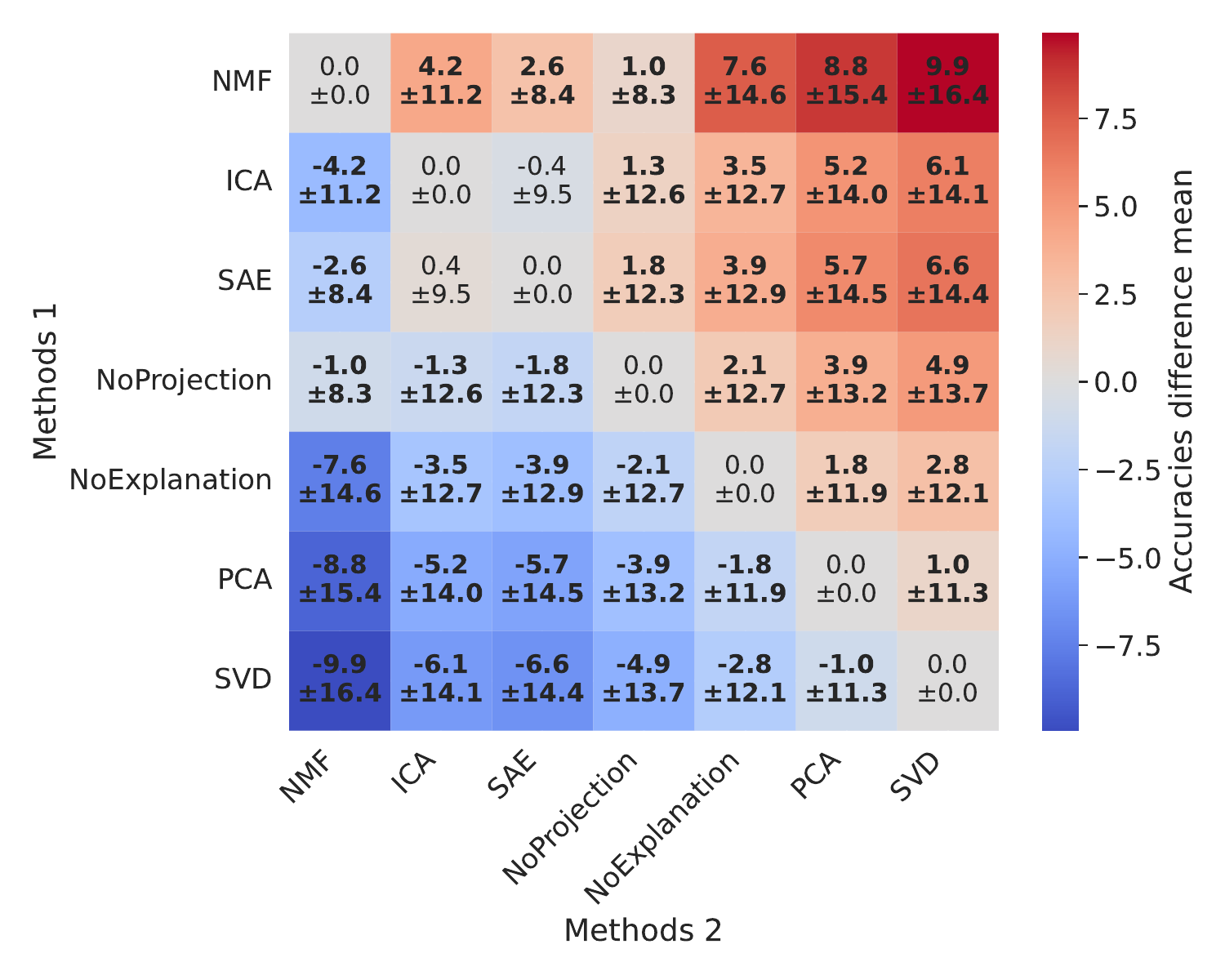}
            \end{subfigure}
            \caption{\textbf{Copy of original ConSim paper plots.} Pairwise win rates (left) and mean score differences with standard deviations (right), using GPT-4o-mini scores on BIOS, Tweet Eval Emotion, IMDB, and Rotten Tomatoes.}
            \label{fig:consim_copy_pairwise}
        \end{figure*}

        To simplify comparison, we copy two figures from the ConSim paper. Fig.~\ref{fig:consim_copy_pairwise} is a screen capture from the original paper, from which we have the accord.
    
\begin{figure*}[tp]
            \centering
            \begin{subfigure}{.49\textwidth}
                \centering
                \includegraphics[width=\linewidth]{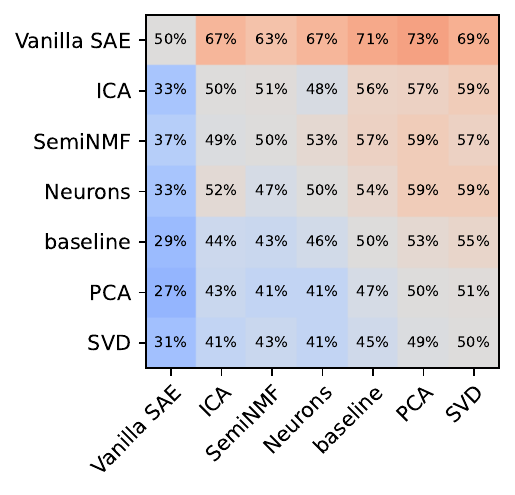}
            \end{subfigure}%
            \begin{subfigure}{.49\textwidth}
                \centering
                \includegraphics[width=\linewidth]{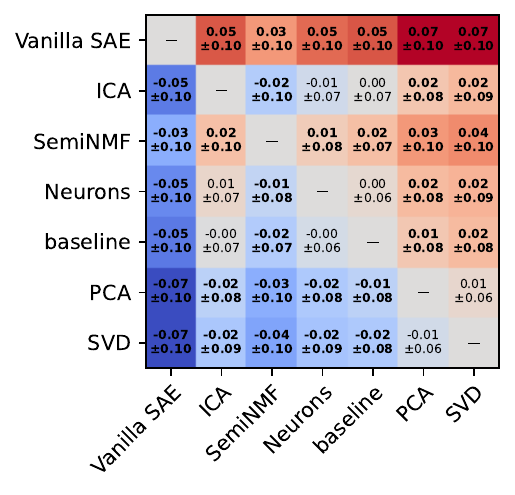}
            \end{subfigure}
            \caption{\textbf{Reconstructed old-ConSim concept-method comparison.} Pairwise win rates (left) and mean score differences with standard deviations (right), using Qwen-3.5-9B scores on BIOS, Emotion, IMDB, and Rotten Tomatoes.}
            \label{fig:old_pairwise}
        \end{figure*}

    \paragraph{Reconstructed old ConSim.}
        The old-format reproduction in Fig.~\ref{fig:old_pairwise} uses the four datasets closest to the original study: BIOS, Emotion, IMDB, and Rotten Tomatoes. It includes the five standard and five anonymized prompt types with TopK interpretations. Vanilla SAE ranks first, followed by ICA and Semi-NMF; PCA and SVD remain at the bottom. Seventeen of the 21 pairwise differences are significant after correction.
    
\begin{figure*}[tp]
            \centering
            \begin{subfigure}{.49\textwidth}
                \centering
                \includegraphics[width=\linewidth]{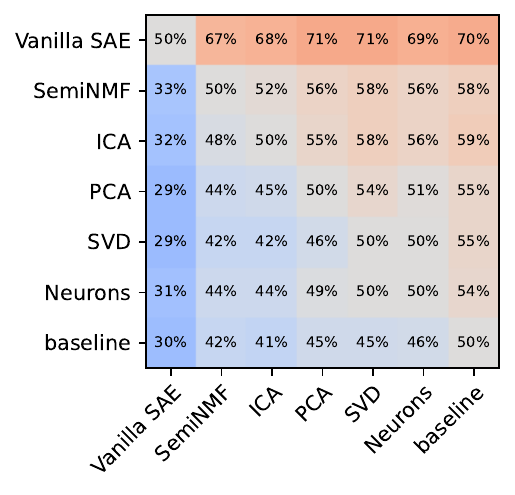}
            \end{subfigure}%
            \begin{subfigure}{.49\textwidth}
                \centering
                \includegraphics[width=\linewidth]{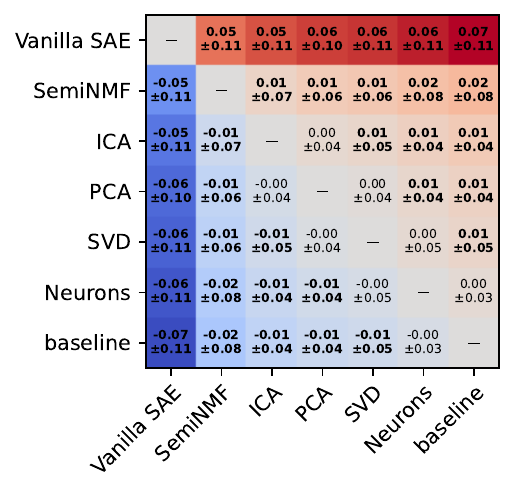}
            \end{subfigure}
            \caption{\textbf{New-ConSim concept-method comparison.} Pairwise win rates (left) and mean score differences with standard deviations (right), using Qwen-3.5-9B scores on all six datasets.}
            \label{fig:new_pairwise}
        \end{figure*}

    \paragraph{New ConSim.}
        Figure~\ref{fig:new_pairwise} applies the same calculation to \texttt{new\_consim} and extends it to all six datasets. Every method pair has all 600 expected matched cells: ten dataset/class-subset settings, ten grouped seeds, and six aligned concept prompt types after matching each concept prompt to its corresponding baseline. Vanilla SAE remains first, followed by Semi-NMF and ICA. Differences among the lower-ranked methods are small despite their ordering. Holm correction retains 17 of the 20 pairwise differences that were significant before correction.
    
\begin{figure*}[tp]
        \centering
        \begin{subfigure}{\textwidth}
            \centering
            \includegraphics[width=\linewidth]{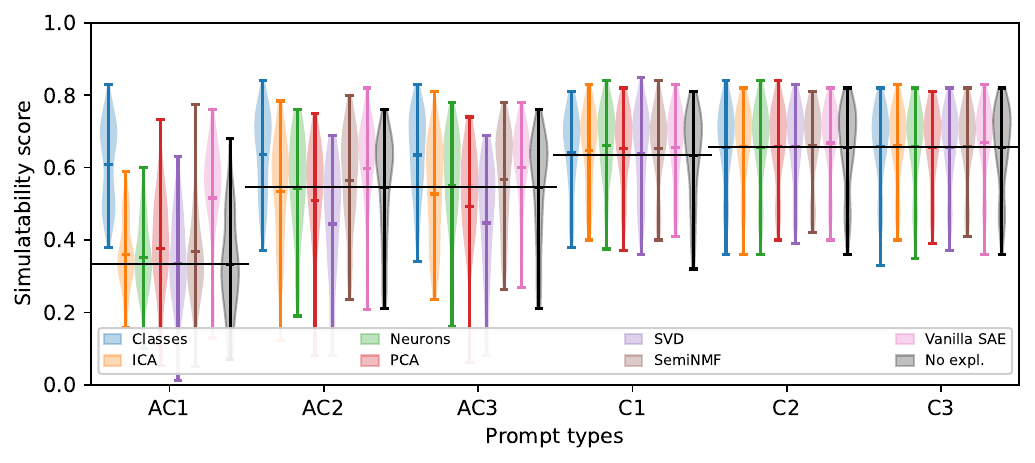}
        \end{subfigure}\\
        \begin{subfigure}{\textwidth}
            \centering
            \includegraphics[width=\linewidth]{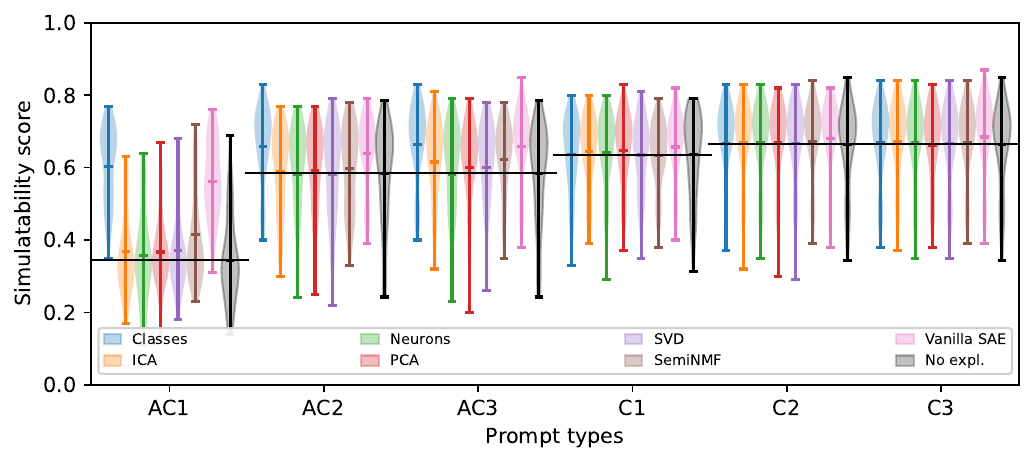}
        \end{subfigure}
        \caption{\textbf{Old- and new-ConSim score distributions.} Qwen-3.5-9B grouped-seed scores for TopK concept configurations under old (top) and new (bottom) prompting. Non-anonymized C1--C3 distributions largely overlap across methods and the matched no-explanation baseline; anonymized AC1--AC3 prompts produce greater separation.}
        \label{fig:old_new_TopK_violins}
    \end{figure*}

    \paragraph{Score distributions.}
        The matrices aggregate small paired differences that can be difficult to assess from rankings alone. Figure~\ref{fig:old_new_TopK_violins} therefore shows the underlying grouped-seed score distributions for the TopK grid, including classes-as-concepts as a diagnostic. Under non-anonymized C1--C3 prompts, method distributions overlap strongly with the matching no-explanation baselines in both formats. Most visible method separation occurs for anonymized AC1--AC3 prompts, where classes-as-concepts can expose the hidden class mapping. This distinction motivates the separate anonymization analysis in Sec.~\ref{sec:anonymized} and App.~\ref{sec:anonymize_pairwise}.
    
\clearpage

% \begin{figure*}[!t]
%         \centering
%         \includegraphics[width=0.1\linewidth]{plots/new_old_diff_datasets.pdf}
%         \caption{\textbf{Dataset-wise old/new prompt-format differences.} Mean paired Qwen-3.5-9B score differences $\texttt{new\_consim}-\texttt{old\_consim}$, with one-standard-deviation error bars. Bars aggregate methods, TopK and LLM interpretations, class subsets, and grouped seeds within each dataset and prompt type.}
%         \label{fig:new_old_diff_datasets}
%     \end{figure*}

    \begin{strip}
        \centering
        \includegraphics[width=\linewidth]{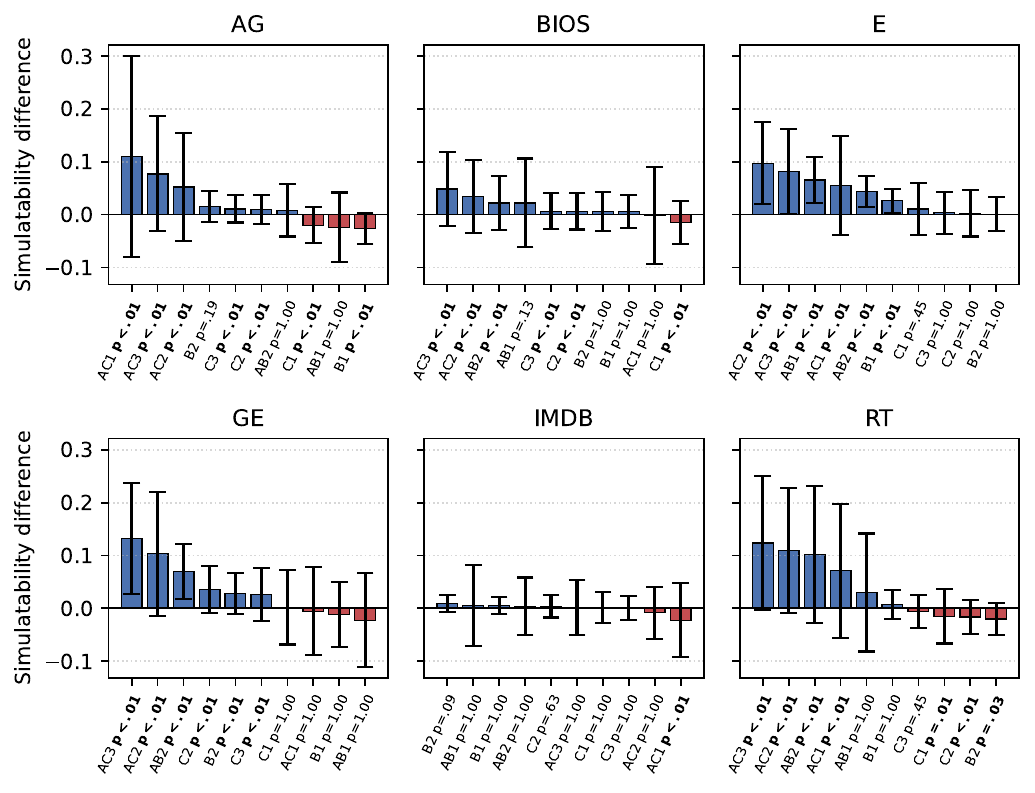}
        \captionof{figure}{\textbf{Dataset-wise old/new prompt-format differences.} Mean paired Qwen-3.5-9B score differences $\texttt{new\_consim}-\texttt{old\_consim}$, with one-standard-deviation error bars. Bars aggregate methods, TopK and LLM interpretations, class subsets, and grouped seeds within each dataset and prompt type.}
        \label{fig:new_old_diff_datasets}
    \end{strip}

\section{Old- and new-ConSim comparison with Qwen-3.5-9B} \label{sec:old_new_comparison_appendix}

    This appendix disaggregates the aggregate old/new comparison in Fig.~\ref{fig:new_old_diff} by dataset. We pair \texttt{old\_consim} and \texttt{new\_consim} scores on every remaining index level: dataset, task model, class subset, grouped seed, method, interpretation, and prompt type. The plotted value is always $\texttt{new\_consim}-\texttt{old\_consim}$, so positive bars favor the new organization. The comparison spans 9,160 grouped seeds exact matches.

    Figure~\ref{fig:new_old_diff_datasets} shows substantial dataset heterogeneity. After Holm correction across all 60 displayed dataset--prompt-type tests, 33 differences are significant: 26 favor the new format and seven favor the old format. Emotion and GoEmotions each have six significantly positive prompt types. IMDB has no significantly positive difference and a significantly negative AC1 difference, while Rotten Tomatoes combines four positive and three negative differences. Thus, reorganizing the prompt usually helps, especially for anonymized concept prompts, but it does not induce a uniform offset across datasets or prompt types.

\clearpage

\begin{strip}
    \centering
    \begin{minipage}{\textwidth}
        \centering
        \captionsetup{type=figure}

        \begin{subfigure}[t]{0.49\textwidth}
            \centering
            \includegraphics[width=\linewidth]{
                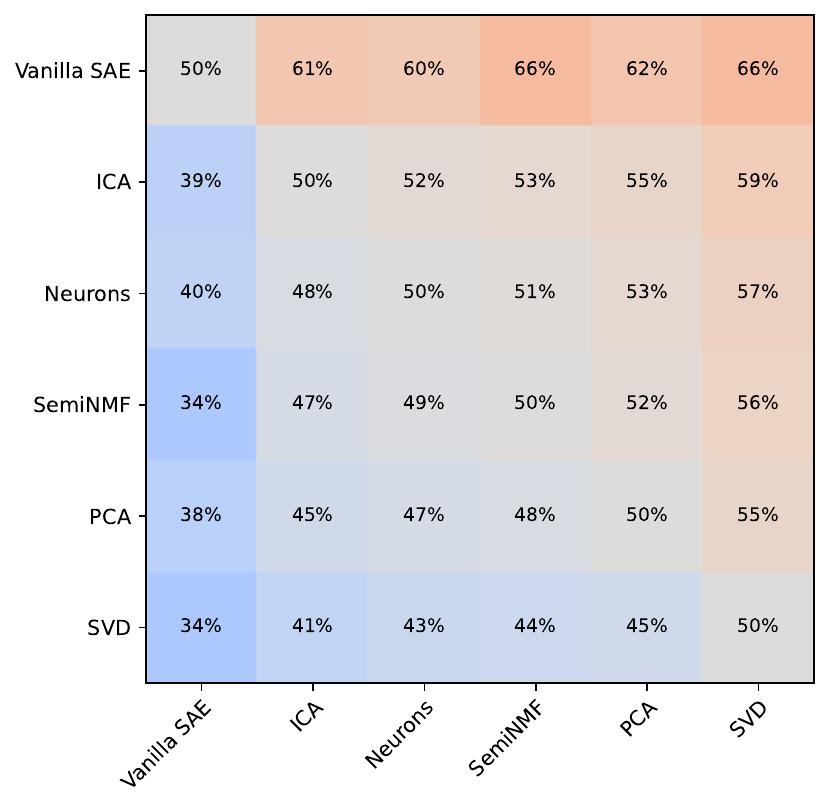
            }
            % Optional:
            % \caption{Pairwise win rates.}
            % \label{fig:concepts_method_winrate_left}
        \end{subfigure}
        \hfill
        \begin{subfigure}[t]{0.49\textwidth}
            \centering
            \includegraphics[width=\linewidth]{
                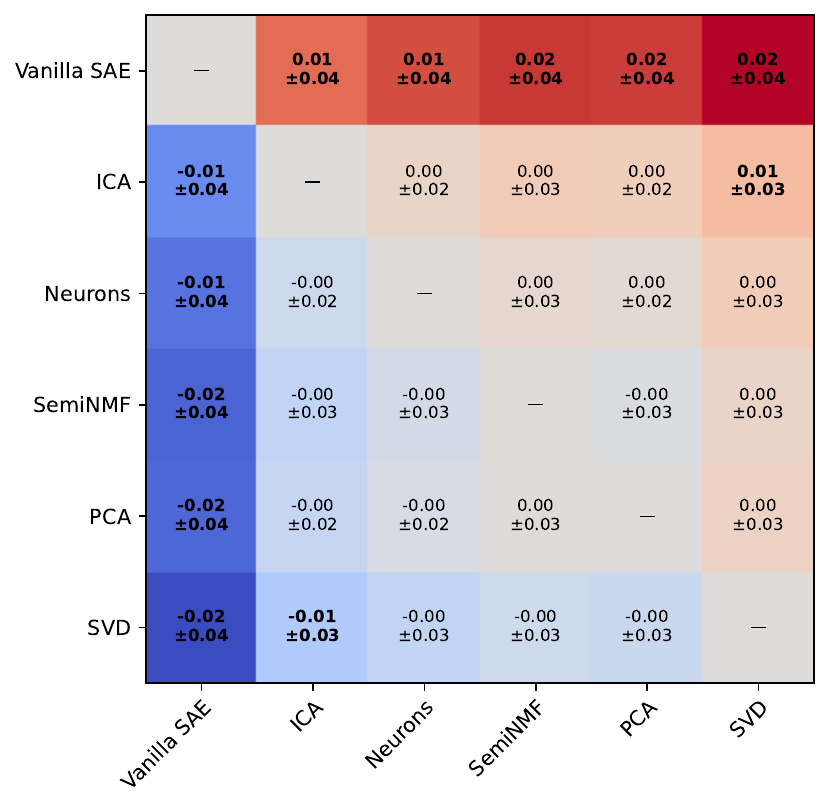
            }
            % Optional:
            % \caption{Mean score differences.}
            % \label{fig:concepts_method_winrate_right}
        \end{subfigure}

        \caption{
            \textbf{Concept-method ranking.} Pairwise win rates (left) and mean score differences with standard deviations (right) for TopK C1--C3 prompts. Every pair contains 300 matched cells.
        }
        \label{fig:concepts_method_winrate}
    \end{minipage}
\end{strip}
\section{Explanation-method rankings with Qwen-3.5-9B}
\label{sec:methods_ranking}

    We rank candidate methods separately within each explanation family using Qwen-3.5-9B scores. All comparisons are non-anonymized: concept methods use \texttt{new\_consim}, TopK interpretations, and C1--C3. Baselines and classes-as-concepts are not ranking candidates. For each method pair, we intersect all remaining experimental keys, count a tie as half a win, and order methods by the same Copeland-style non-loss count used in the concept analysis. Difference matrices report paired means and standard deviations; bold cells survive Holm correction within that explanation family.
    
    \paragraph{Concepts.}
        The 6 compared decompositions are PCA \cite{pearson1901liii,hotelling1992relations}, SVD \cite{eckart1936approximation}, \cite{ans1985architectures,hyvarinen2000independent}, Semi-NMF \cite{ding2010convex}, and SAEs \cite{ng2011sparse,makhzani2013k,domingos2015master}.
        
        Figure~\ref{fig:concepts_method_winrate} compares six methods on 300 matched cells per pair. Vanilla SAE ranks first, with win rates of 60--66\% against every contender and mean advantages of 0.013--0.018. All five of these differences remain significant after Holm correction over the 15 concept-method pairs. Six pairs are significant overall, the additional pair being ICA over SVD. We therefore retain Vanilla SAE as the representative concept method.

        \begin{table}[t]
            \centering
            \footnotesize
            \setlength{\tabcolsep}{3pt}
            \begin{tabular}{p{.18\linewidth}p{.18\linewidth}p{.50\linewidth}}
                \toprule
                \textbf{Class} & \textbf{Method} & \textbf{Displayed global descriptor} \\
                \midrule
                dentist & SVD & Scholarship, scholarship, Ph.D., Professor, Faculty (opposed) \\
                dentist & Vanilla SAE & D.D.S., Dentistry, teeth, board-certified, registry (supportive) \\
                physician & SVD & Scholarship, scholarship, Ph.D., Professor, Faculty (supportive) \\
                physician & Vanilla SAE & Hematology/Oncology, hospital/clinic, Hematology, patient-centered, preventative (supportive) \\
                surgeon & SVD & None displayed \\
                surgeon & Vanilla SAE & medicine/transfusion-free, hospital/clinic, medical-surgical, Dr, therapeutic (supportive) \\
                \bottomrule
            \end{tabular}
            \caption{\textbf{Qualitative BIOS TopK concept examples.} Representative words or phrases from the first displayed globally important concept for each class in the \{dentist, physician, surgeon\} subset. Descriptors and directions are reproduced from the non-anonymized \texttt{new\_consim} prompts.}
            \label{tab:qualitative_concept_examples}
        \end{table}

    \paragraph{Attributions.}
        The 10 compared attribution methods are: Saliency \cite{simonyan2014deep}, Occlusion \cite{zeiler2014visualizing}, Lime \cite{ribeiro2016should}, KernelShap \cite{lundberg2017unified}, SmoothGrad \cite{smilkov2017smoothgrad}, Integrated Gradients \cite{sundararajan2017axiomatic}, SquareGrad \cite{adebayo2018sanity}, VarGrad \cite{hooker2018benchmark}, and Sobol \cite{fel2021sobol}. Note that we use the gradient-input \cite{shrikumar2017learning} version of all gradient-based methods.

        \begin{figure*}[!t]
            \centering
            \begin{subfigure}{.49\textwidth}
                \centering
                \includegraphics[width=\linewidth]{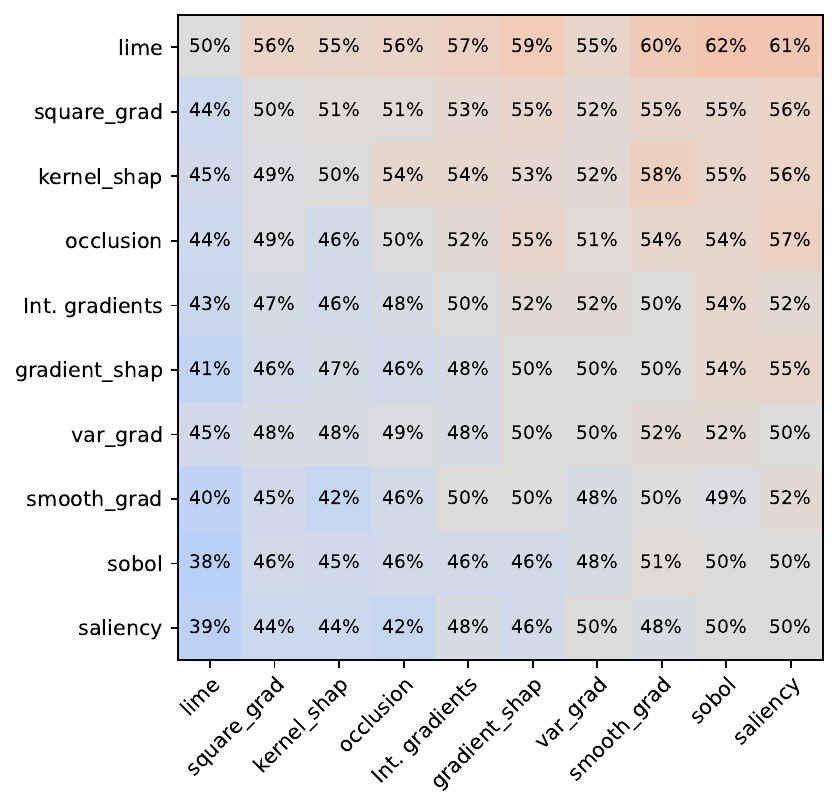}
            \end{subfigure}%
            \begin{subfigure}{.49\textwidth}
                \centering
                \includegraphics[width=\linewidth]{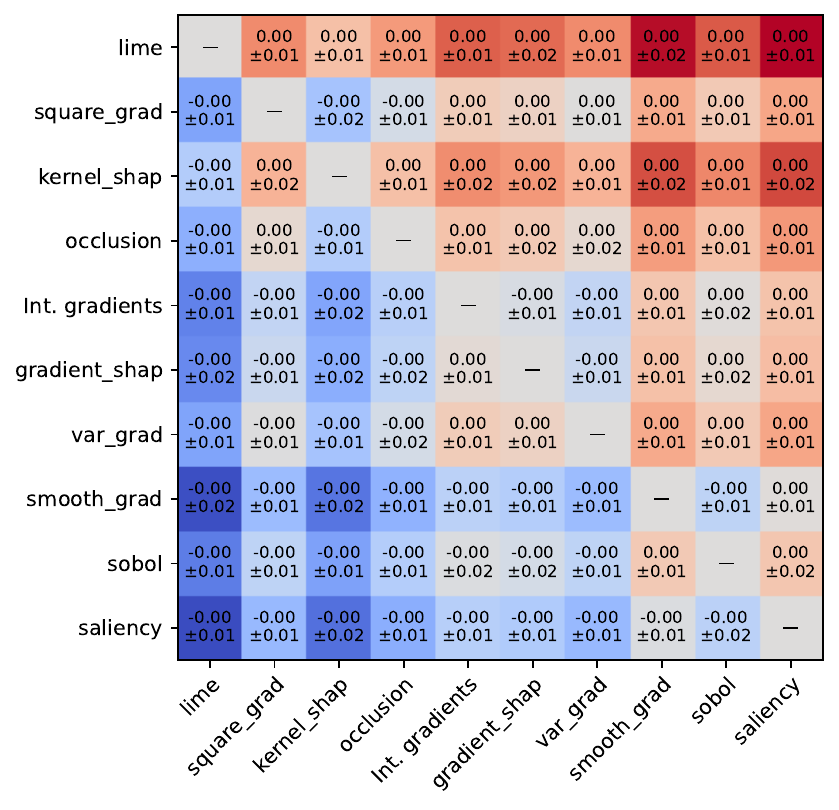}
            \end{subfigure}
            \caption{\textbf{Attribution-method ranking.} Pairwise win rates (left) and mean score differences with standard deviations (right) for A1 prompts. Pairs contain 100 matched cells, except kernel-SHAP pairs, which contain 90. No comparison is significant after Holm correction.}
            \label{fig:attributions_method_winrate}
        \end{figure*}
    
        Figure~\ref{fig:attributions_method_winrate} compares ten methods. LIME ranks first and wins 55--62\% of its matched comparisons, but its mean advantages are only 0.001--0.004. None of the 45 attribution-method differences remains significant after Holm correction. LIME is therefore a deterministic first-ranked representative, not evidence of a reliably superior attribution method.
    
    \paragraph{Rationales.}
        Figure~\ref{fig:rationales_method_winrate} compares the two available rationale generators on 100 matched cells. Qwen3.5-2B has a 57\% win rate over Llama-3.2-3B-Instruct, but the mean difference is only 0.0017 and is not significant ($p=.38$). We use Qwen3.5-2B as the representative while treating the two generators as statistically indistinguishable under this metric. We note that a Qwen model judged the other Qwen model, so the comparison is not fully fair. Nonetheless, due to the score difference, we do not expect it to impact the conclusions in any way.

    \begin{center}
        \begin{minipage}{.9\linewidth}
        \centering
        \includegraphics[width=\linewidth]{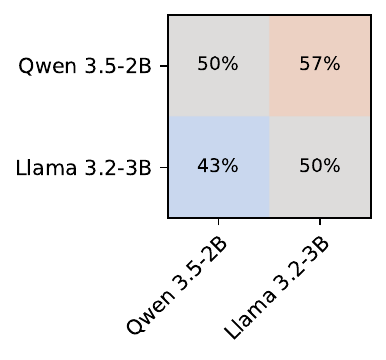}\\
        \includegraphics[width=\linewidth]{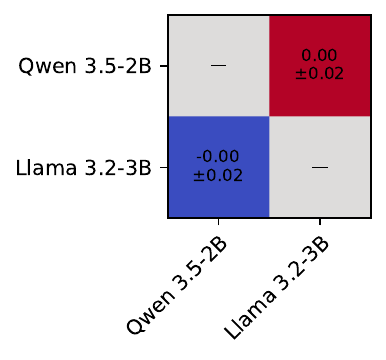}
        \captionof{figure}{\textbf{Rationale-generator ranking.} Pairwise win rates (left) and mean score differences with standard deviations (right) for R1 prompts. The single comparison contains 100 matched cells and is not significant.}
        \label{fig:rationales_method_winrate}
        \end{minipage}
    \end{center}

\clearpage
\section{Explanation-family score distributions}
\label{sec:family_score_distributions}

\begin{figure*}[t]
        \centering
        \begin{subfigure}{0.78\textwidth}
            \centering
            \includegraphics[width=\linewidth]{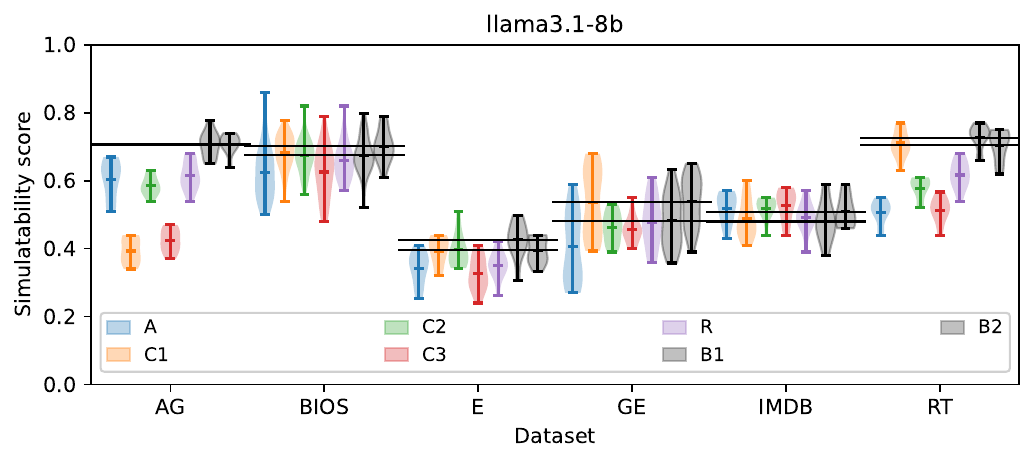}
        \end{subfigure}\\
        \begin{subfigure}{0.78\textwidth}
            \centering
            \includegraphics[width=\linewidth]{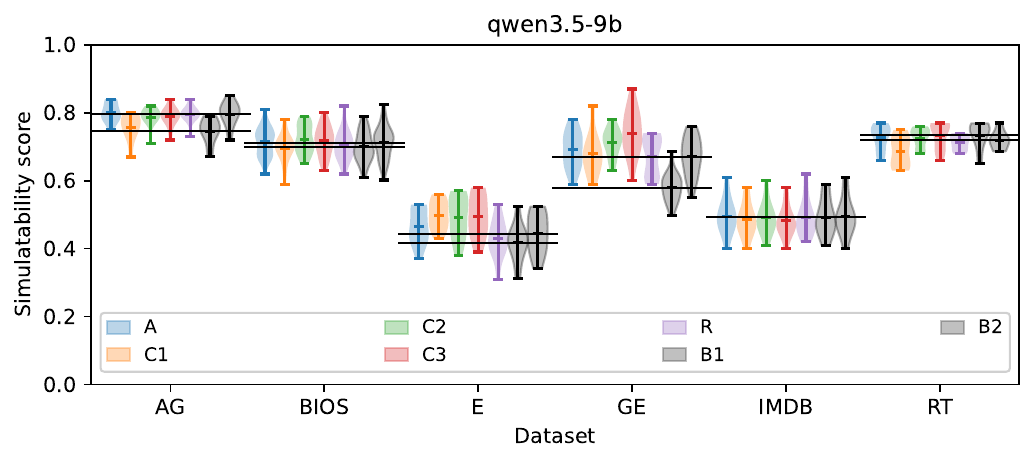}
        \end{subfigure}\\
        \begin{subfigure}{0.78\textwidth}
            \centering
            \includegraphics[width=\linewidth]{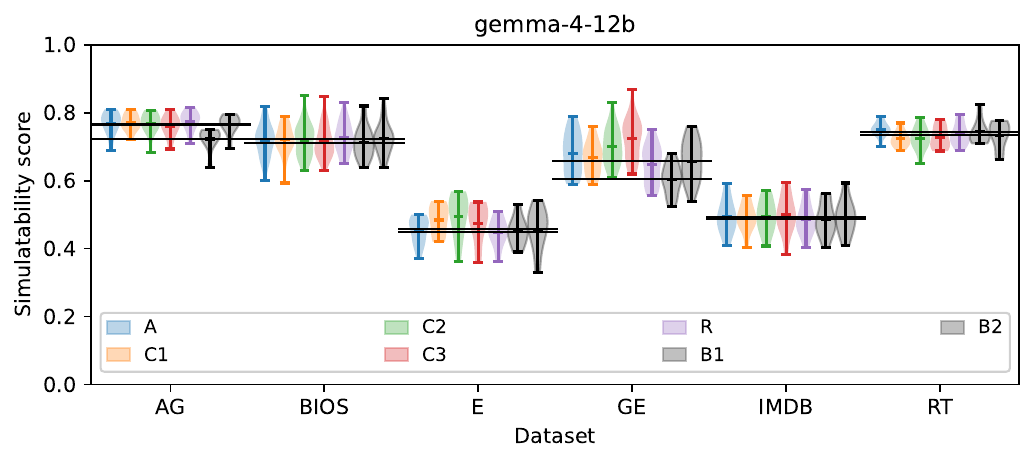}
        \end{subfigure}\\
        \begin{subfigure}{0.78\textwidth}
            \centering
            \includegraphics[width=\linewidth]{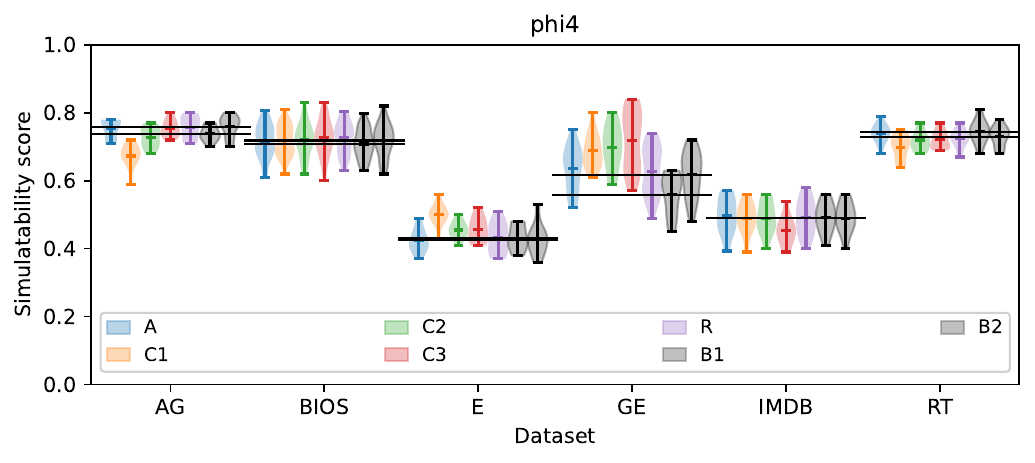}
        \end{subfigure}
        \caption{\textbf{Selected explanation-family score distributions.} Grouped-seed scores by dataset for Llama-3.1-8B, Qwen-3.5-9B, Gemma-4-12B, and Phi-4 judges. A is LIME, C1--C3 are TopK Vanilla SAE prompts, and R uses Qwen-3.5-2B rationales; B1 and B2 are the matching no-explanation baselines. Horizontal black lines mark the corresponding baseline means within each dataset.}
        \label{fig:families_violins_by_judge}
    \end{figure*}

    Figure~\ref{fig:families_violins_by_judge} shows the grouped-seed score distributions for the representatives selected in App.~\ref{sec:methods_ranking}: Vanilla SAE concepts (C1--C3), LIME attributions (A), and Qwen3.5-2B rationales (R). B1 is the matched baseline for C1, while B2 is the matched baseline for C2, C3, A, and R. Baseline copies are first averaged within each explanation-family specification and then averaged equally across concepts, rationales, and attributions. This prevents repeated concept metadata configurations from receiving extra weight.
    
    Each displayed series contains 100 grouped cells: ten grouped seeds for each of ten dataset/class-subset settings. These pooled distributions are not dataset-balanced because the four BIOS subsets contribute 40 cells and the two GoEmotions subsets contribute 20, while each other dataset contributes ten. All grouped cells are present.
    
    Qwen, Phi, and Gemma show broad overlap, with pooled prompt-type means concentrated around 0.64--0.68. Their concept improvements over matching baselines are generally only one to three score points and vary substantially by dataset. Rationales do not reliably improve over B2 for any of these three judges, and attributions improve significantly only for Qwen. Llama behaves differently: B2 has the highest pooled mean, while A, C2, C3, and R are lower. Thus, no explanation family has a stable advantage across judges and datasets; App.~\ref{sec:prediction_exact_agreement} quantifies these paired differences.
    
    Llama being the least performant model of the bunch, we could dismiss its results as a failure of the judge. With this, results are much more aligned.
    
\clearpage
\begin{strip}
    \centering
    \begin{minipage}{\textwidth}
    \centering
    \captionsetup{type=figure}
    \begin{subfigure}{0.4\textwidth}
        \centering
        \includegraphics[width=\linewidth]{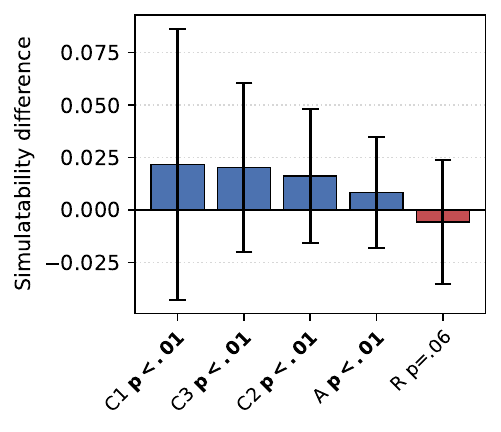}
        \caption{Qwen-3.5-9B}
    \end{subfigure}
    \begin{subfigure}{0.4\textwidth}
        \centering
        \includegraphics[width=\linewidth]{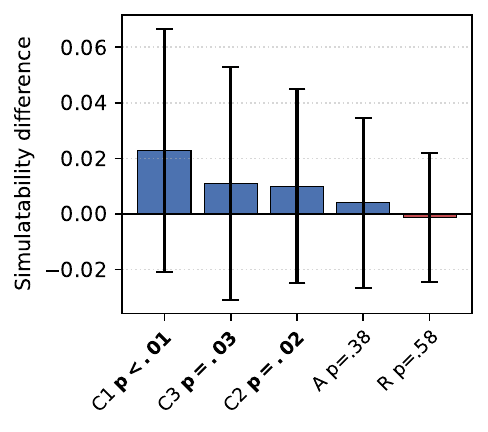}
        \caption{Gemma-4-12B}
    \end{subfigure}\\
    \begin{subfigure}{0.4\textwidth}
        \centering
        \includegraphics[width=\linewidth]{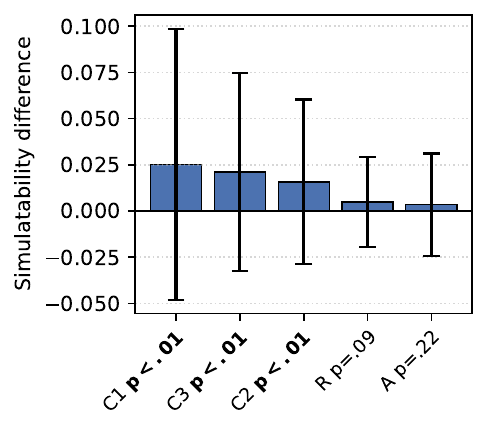}
        \caption{Phi-4}
    \end{subfigure}
    \begin{subfigure}{0.4\textwidth}
        \centering
        \includegraphics[width=\linewidth]{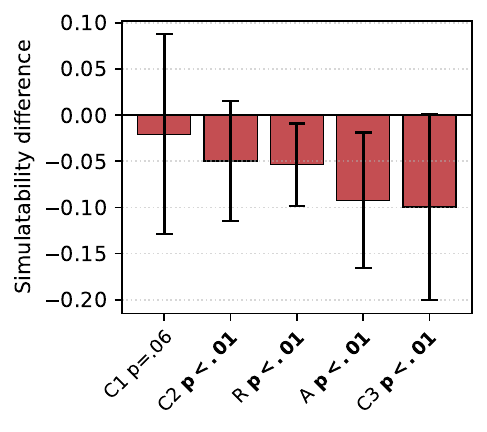}
        \caption{Llama-3.1-8B}
    \end{subfigure}
    \caption{\textbf{Matched explanation--baseline differences across LLM simulators.} Bars report mean paired simulatability-score differences, with one-standard-deviation error bars. C1 is compared with B1; C2, C3, attribution (A), and rationale (R) are compared with B2. Positive values indicate higher scores when explanations are included. Displayed $p$-values are obtained from paired Student $t$-tests and Holm-corrected.}
    \vspace{-0.5em}
    \label{fig:families_differences_by_judge}
    \end{minipage}
\end{strip}
\section{Matched explanation--baseline differences across LLM simulators}
\label{sec:llm_judges_barplots}

    Figure~\ref{fig:families_violins_by_judge} compares each explanation prompt with its matched no-explanation baseline: C1 is compared with B1, while C2, C3, A, and R are compared with B2. Comparisons are matched on dataset, task model, class subset, and grouped seed, yielding 100 paired grouped-seed observations for each contrast and LLM simulator. Bars report the mean paired simulatability-score difference, with one standard deviation. The displayed $p$-values come from two-sided paired Student $t$-tests with Holm correction across the five comparisons for each LLM simulator. Subset settings are weighted equally, so BIOS and GoEmotions receive greater aggregate weight than single-setting datasets.
    
    \paragraph{Qwen-3.5-9B.}
        C1, C2, C3, and A significantly outperform their matched baselines. R is slightly below B2, but the difference is not significant.
    
    \paragraph{Gemma-4-12B.}
        Gemma-4-12B shows the same ordering as Qwen-3.5-9B. C1 and C3 significantly outperform their baselines, while the smaller differences for C2, A, and R are not significant.
    
    \paragraph{Phi-4.}
        All five differences are positive. C1, C2, and C3 are significant, while A and R are not.
    
    \paragraph{Llama-3.1-8B.}
        All five differences are negative. C2, C3, A, and R are significantly below B2, while C1 is not significantly different from B1.
    
    \paragraph{Cross-LLM simulator pattern.}
        Qwen-3.5-9B, Gemma-4-12B, and Phi-4 show a similar ordering: concept prompts provide the largest gains, while attribution and rationale provide smaller or null gains. Llama-3.1-8B is the main exception.
    
    These results support using multiple LLM simulators and reporting effect sizes rather than relying only on statistical significance.

\clearpage
\begin{strip}
    \centering
    \begin{minipage}{\textwidth}
    \centering
    \captionsetup{type=figure}
    \begin{subfigure}{.48\textwidth}
        \centering
        \includegraphics[width=\linewidth]{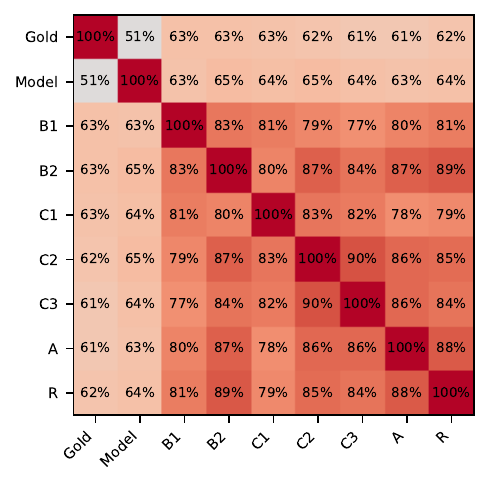}
    \end{subfigure}\hfill
    \begin{subfigure}{.48\textwidth}
        \centering
        \includegraphics[width=\linewidth]{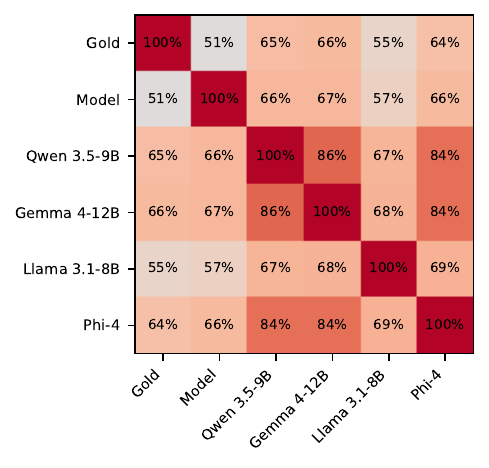}
    \end{subfigure}
    \caption{\textbf{Aggregate prediction exact agreement.} Mean pairwise exact agreement among prompt types, gold labels, and task model predictions (left), and among LLM simulator after stacking prompt types (right). Agreements are averaged with equal weight; missing predictions are handled pairwise.}
    \label{fig:prediction_exact_agreement_summaries}
    \end{minipage}
\end{strip}
\section{Prompt-prediction and judge exact agreement}
\label{sec:prediction_exact_agreement}

    This appendix analyzes the sample-level class predictions. We retain the non-anonymized representatives selected in App.~\ref{sec:methods_ranking}: Vanilla SAE with TopK interpretations for C1--C3, LIME for A, and Qwen3.5-2B for R, together with B1 and B2. To avoid stitching together disagreeing duplicate baseline generations, B1 and B2 use the attribution specification as one declared baseline source for every judge. The analysis covers all six datasets, including GoEmotions and BIOS.
    
    Exact agreement is the percentage of pairwise-complete predictions assigned the same class label. It is computed separately within each setting and then averaged with equal weight. The overall prompt matrix uses 40 judge/dataset/class-subset settings.
    
    Figure~\ref{fig:prediction_exact_agreement_summaries} summarizes the two views. Off-diagonal prompt-type agreements range from 77\% to 90\%. This includes the B1 baseline, which has neither learning examples nor explanations, showing that much of the LLM simulator prediction remains unchanged across prompt types.
    
\begin{figure*}[t]
        \centering
        \includegraphics[width=\linewidth,height=.82\textheight,keepaspectratio]{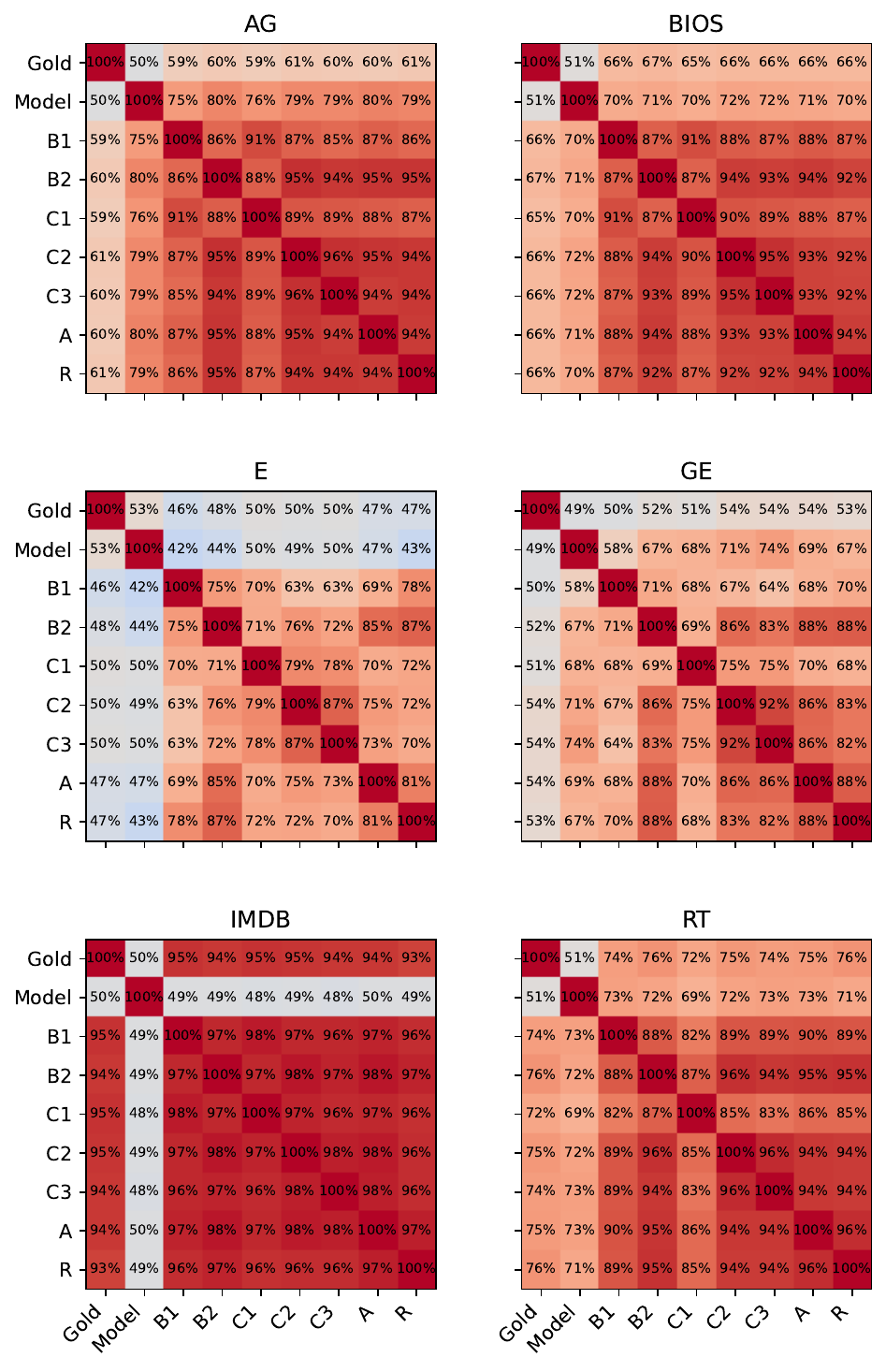}
        \caption{\textbf{Qwen-3.5-9B prediction exact agreement by dataset.} Exact agreement among gold labels, task model predictions, and selected prompt types, averaged equally over class subsets within each dataset.}
        \label{fig:prediction_exact_agreement_qwen}
    \end{figure*}

\begin{figure*}[t]
        \centering
        \includegraphics[width=\linewidth,height=.82\textheight,keepaspectratio]{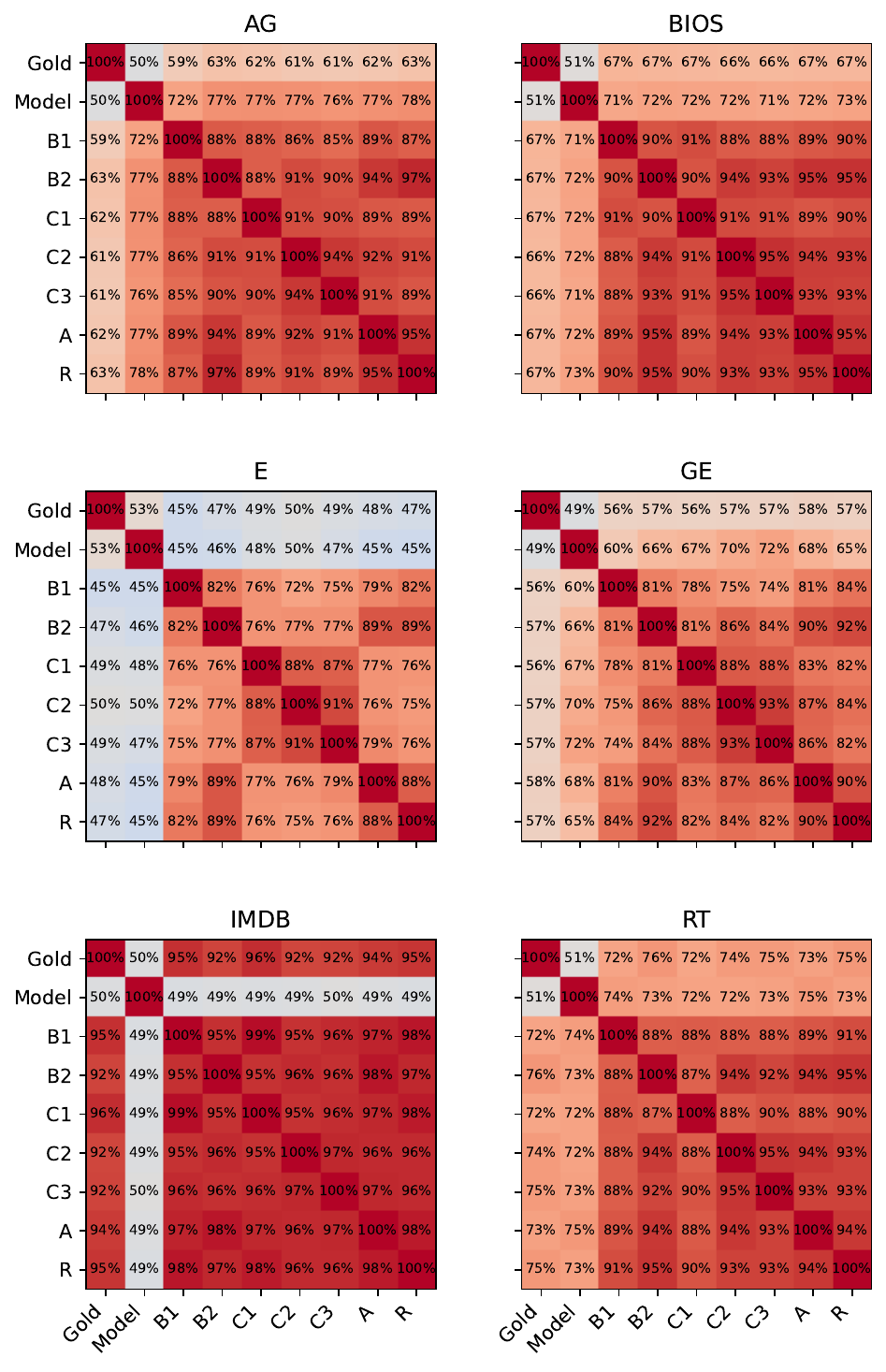}
        \caption{\textbf{Gemma-4-12B prediction exact agreement by dataset.} Exact agreement among gold labels, task model predictions, and selected prompt types, averaged equally over class subsets within each dataset.}
        \label{fig:prediction_exact_agreement_gemma}
    \end{figure*}

\begin{figure*}[t]
        \centering
        \includegraphics[width=\linewidth,height=.82\textheight,keepaspectratio]{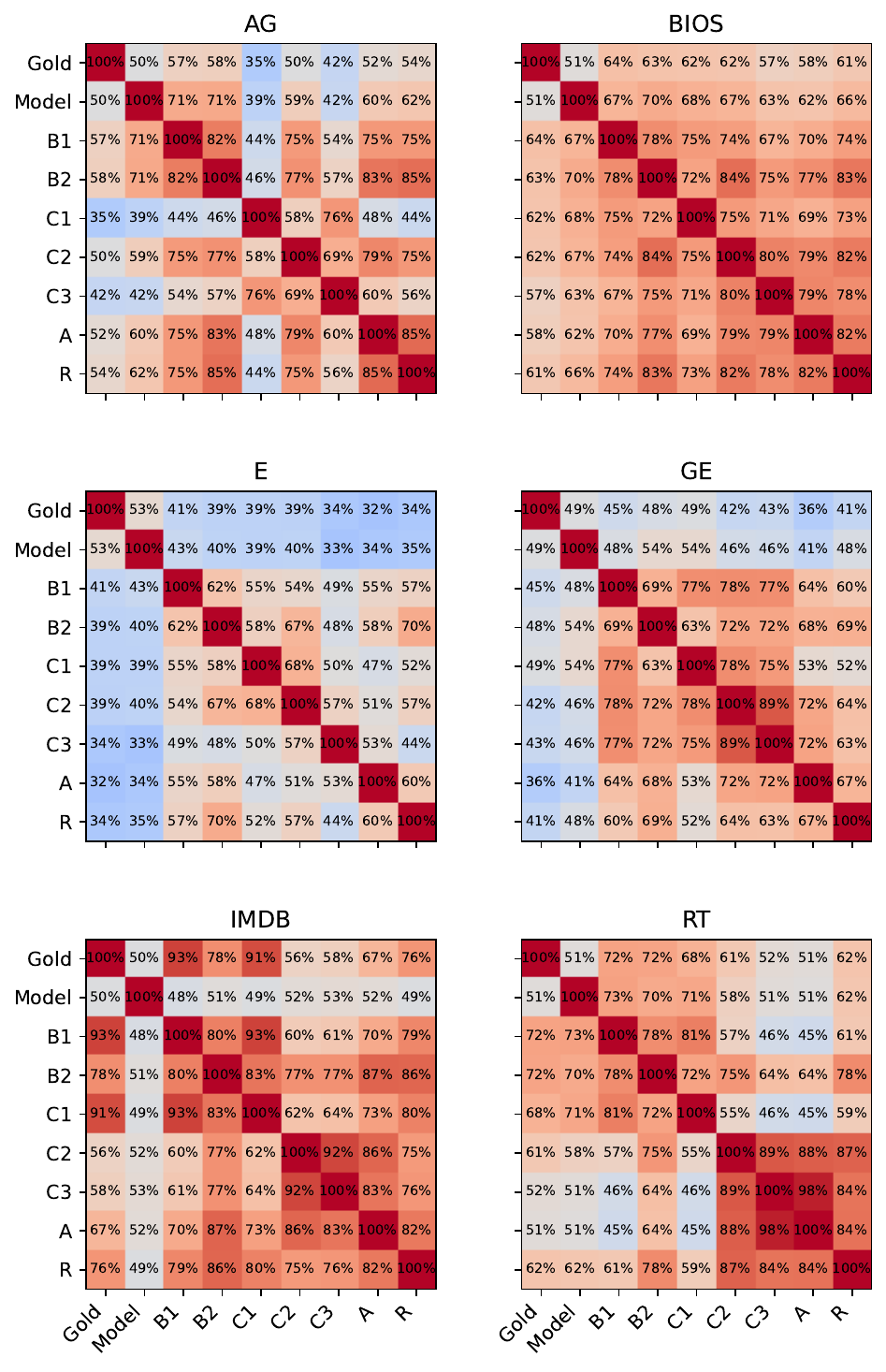}
        \caption{\textbf{Llama-3.1-8B prediction exact agreement by dataset.} Exact agreement among gold labels, task model predictions, and selected prompt types, averaged equally over class subsets within each dataset.}
        \label{fig:prediction_exact_agreement_llama}
    \end{figure*}

\begin{figure*}[t]
        \centering
        \includegraphics[width=\linewidth,height=.82\textheight,keepaspectratio]{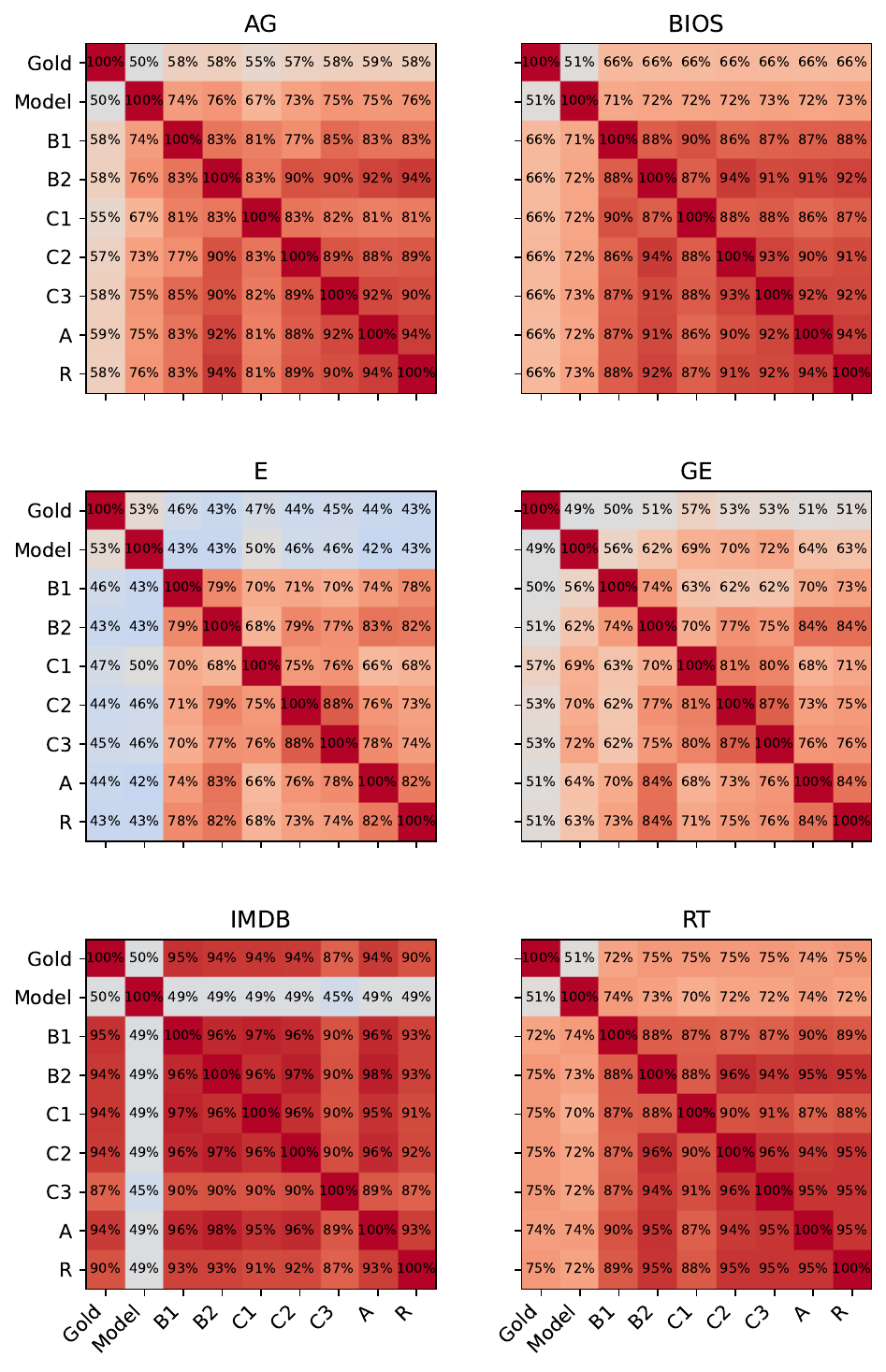}
        \caption{\textbf{Phi-4 prediction exact agreement by dataset.} Exact agreement among gold labels, task model predictions, and selected prompt types, averaged equally over class subsets within each dataset.}
        \label{fig:prediction_exact_agreement_phi}
    \end{figure*}

    Figures~\ref{fig:prediction_exact_agreement_qwen}--\ref{fig:prediction_exact_agreement_phi} disaggregate the prompt matrix by judge and dataset. Agreement varies substantially by dataset and judge. On AG News and BIOS, Qwen-3.5-9B, Gemma-4-12B, and Phi-4 generally retain moderate-to-high agreement with the task model, whereas Llama-3.1-8B is more variable. On IMDB, prompt/task model agreement is only 45--53\% for every judge, close to the task model's 50\% agreement with gold in this deliberately balanced correct/error sample. Agreement with gold is much larger: B1 and C1 exceed 90\% for every judge. The simulators mostly follow the sentiment labels, providing direct evidence of task solving instead of task model simulation in that setting; the diverse LLM simulators agree on this pattern.
    
    These matrices are descriptive, not significance tests, and high prompt-to-prompt exact agreement does not by itself prove that explanations are unused. Combined with the byte-identical baselines, small explanation--baseline score differences, and the IMDB failure mode, it nevertheless supports the shortcut hypothesis: much of the simulator prediction is determined by the text classification task and judge prior rather than explanation-specific information.
    
\clearpage
\section{Anonymized and non-anonymized new-ConSim comparisons}
\label{sec:anonymize_pairwise}

    This appendix isolates the effect of class anonymization on the Qwen-3.5-9B concept-method ranking. Both matrices use \texttt{new\_consim}, TopK interpretations, all six datasets, and the paper grid of Semi-NMF, ICA, PCA, SVD, Vanilla SAE, neurons-as-concepts, classes-as-concepts, and the matched no-explanation baseline. As in App.~\ref{sec:consim_reproduction_appendix}, exact duplicates are averaged and every five seeds form one grouped-seed observation. Each method pair has 300 matched cells: ten dataset/class-subset settings, ten grouped seeds, and three prompt types. Holm correction is applied separately across the 28 method pairs in each matrix.
    
\begin{figure*}[t]
        \centering
        \begin{subfigure}{.49\textwidth}
            \centering
            \includegraphics[width=\linewidth]{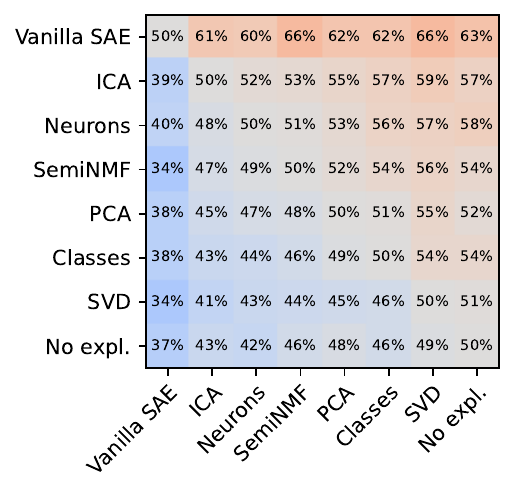}
        \end{subfigure}%
        \begin{subfigure}{.49\textwidth}
            \centering
            \includegraphics[width=\linewidth]{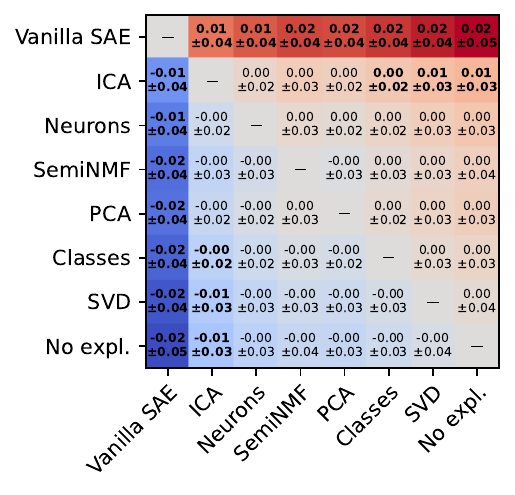}
        \end{subfigure}
        \caption{\textbf{Non-anonymized new-ConSim concept comparison.} Pairwise win rates (left) and mean score differences with standard deviations (right) for C1--C3. Every pair contains 300 matched cells.}
        \label{fig:new_TopK_non_anon_pairwise}
    \end{figure*}

    \paragraph{Non-anonymized prompts.}
        Figure~\ref{fig:new_TopK_non_anon_pairwise} aggregates C1--C3. Vanilla SAE ranks first, but the mean differences are small: every displayed gap rounds to at most 0.02. 10 of 28 pairwise differences are significant after correction. classes-as-concepts sits near the lower middle of the ranking and differs from the no-explanation baseline by only 0.002 on average, which is not significant. Thus, its direct use of class names as concept descriptions provides no material advantage when the prompt already reveals them.
    
\begin{figure*}[t]
        \centering
        \begin{subfigure}{.49\textwidth}
            \centering
            \includegraphics[width=\linewidth]{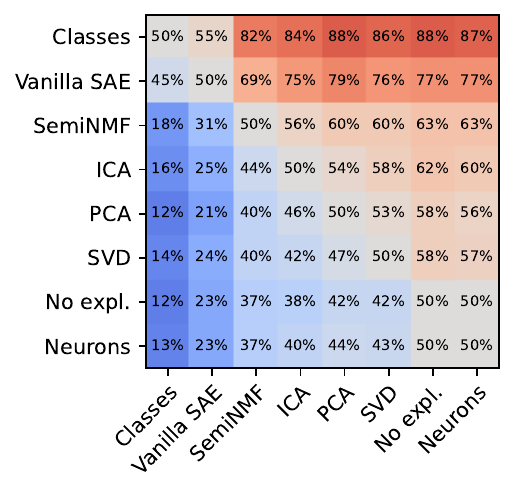}
        \end{subfigure}%
        \begin{subfigure}{.49\textwidth}
            \centering
            \includegraphics[width=\linewidth]{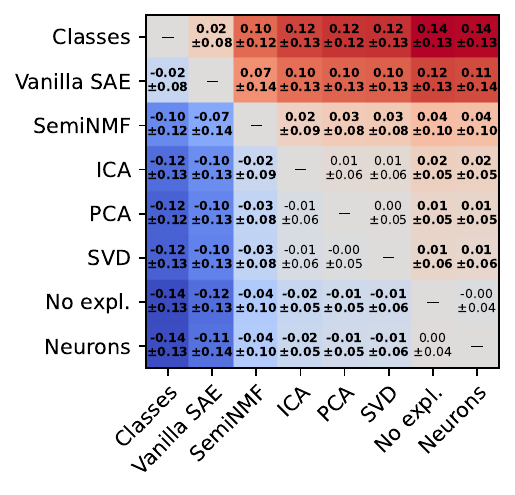}
        \end{subfigure}
        \caption{\textbf{Anonymized new-ConSim concept comparison.} Pairwise win rates (left) and mean score differences with standard deviations (right) for AC1--AC3. Every pair contains 300 matched cells. Classes-as-concepts ranks first because its concept descriptions reveal the hidden class mapping.}
        \label{fig:new_TopK_anon_pairwise}
    \end{figure*}

    \paragraph{Anonymized prompts.}
        Figure~\ref{fig:new_TopK_anon_pairwise} instead aggregates AC1--AC3. Classes-as-concepts now ranks first and beats every other contender after correction. Its mean advantage ranges from 0.022 over Vanilla SAE to 0.138 over the no-explanation baseline, and its pairwise win rates range from 55\% to 88\% against the retained contenders. Overall, 24 of 28 method pairs are significant. The explanation instead acts as a key for translating hidden \texttt{Class\_i} identifiers back to class names. The reversal therefore demonstrates that anonymized simulatability can reward label leakage rather than explanation quality.
    
\end{document}